\documentclass[runningheads]{llncs}

\usepackage{eccv}

\usepackage{booktabs} 
\usepackage{adjustbox}
\usepackage{tikz}
\usepackage[utf8]{inputenc}
\usepackage{pgfplots}
\usepgfplotslibrary{groupplots}
\usepackage{svg}
\usetikzlibrary{patterns}
\usepackage{bm}
\usepackage{multirow}
\usepackage{pifont}
\usepackage{eccvabbrv}
\usepackage{graphicx}
\usepackage{booktabs}
\usepackage{caption}
\usepackage[accsupp]{axessibility}  
\usepackage{ulem}
\usepackage{makecell}
\usepackage{siunitx}

\newcommand{\cmark}{\ding{51}}%
\newcommand{\xmark}{\ding{55}}%

\usepackage{hyperref}

\usepackage{orcidlink}

\pgfplotsset{compat=1.18}

\begin{document}

\title{VL4AD: Vision-Language Models Improve
Pixel-wise Anomaly Detection} 

\titlerunning{VL4AD: Vision-Language Models Improve
Pixel-wise Anomaly Detection}

\author{Liangyu Zhong\inst{1,2}
\orcidlink{0009-0000-4286-9095}
\and
Joachim Sicking\inst{1}
\orcidlink{0000-0003-1741-2338} 
\and \\
Fabian Hueger\inst{1}
\orcidlink{0000-0002-4861-3178} 
\and
\mbox{Hanno Gottschalk}\inst{2}
\orcidlink{0000-0003-2167-2028}}

\authorrunning{Zhong et al.}

\institute{CARIAD SE, Germany\\ \email{\{first.last\}@cariad.technology} 
\vspace{2mm}
\and Institute of Mathematics, Technical University Berlin, Germany\\
\email{gottschalk@math.tu-berlin.de}}

\maketitle


\begin{abstract}
Semantic segmentation networks have achieved significant success under the assumption of independent and identically distributed data. However, these networks often struggle to detect anomalies from unknown semantic classes due to the limited set of visual concepts they are typically trained on. To address this issue, anomaly segmentation often involves fine-tuning on outlier samples, necessitating additional efforts for data collection, labeling, and model retraining. Seeking to avoid this cumbersome work, we take a different approach and propose to incorporate \textit{vision-language} (VL) encoders into existing anomaly detectors to leverage the semantically broad VL pre-training for improved outlier awareness. Additionally, we propose a new scoring function that enables data- and training-free outlier supervision via textual prompts. The resulting \textit{VL4AD} model, which includes max-logit prompt ensembling and a class-merging strategy, achieves competitive performance on widely used benchmark datasets, thereby demonstrating the potential of vision-language models for pixel-wise anomaly detection.
\end{abstract}


\section{Introduction}
\label{sec:intro}
\begin{figure}
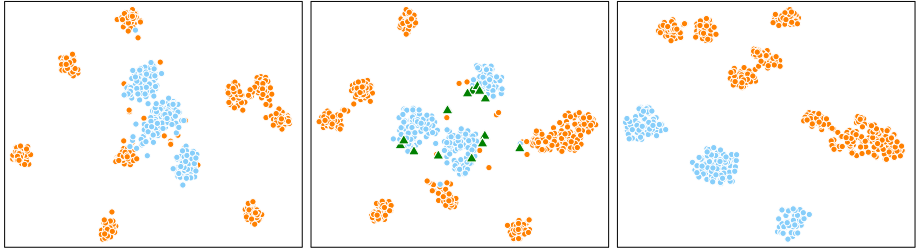

\begin{tabular}{ccc}
  \adjustbox{width=0.325\textwidth}{\fbox{\input{images/fig1/basic.tex}}}
  \hfill
  \adjustbox{width=0.325\textwidth}{\fbox{\input{images/fig1/oe_new.tex}}}
  \hfill
  \adjustbox{width=0.325\textwidth}{\fbox{\input{images/fig1/clip.tex}}}
  \\
\end{tabular}
\caption{\textbf{Showcasing the favorable ID-OOD data separation of a CLIP \cite{Radford2021ICML} image encoder (right) compared to the backbones of vision-only ResNet50 networks \cite{He2016CVPR} (left, middle)}. We use t-SNE \cite{Maaten2008JMLR} to visualize the embedding vectors of images from ImageNet-200 \cite{zhang2023arxiv} (orange points) and OOD samples from the NINCO dataset \cite{bitterwolf2023ICML} (light blue points). OOD samples used for fine-tuning the ResNet50 model (middle) are shown as dark green triangles.}
\label{fig:teaser}
\end{figure}

Recent advances in deep neural networks (DNNs) have led to significant improvements in semantic segmentation tasks for urban driving scenes \cite{Wang2021CVPR, Cheng2022CVPR, Cordts2016CVPR, hümmer2023arxiv}, especially when the semantic classes of training and testing are well aligned \cite{Nguyen2015CVPR}. In real-world situations, however, unexpected object types, that were not part of the training data, appear during operations due to long-tailed class distributions. Examples include wild animals on roads or objects falling from cars on highways. Existing semantic segmentation networks often fail to detect such objects, leading to unreliable predictions that could result in collisions and \mbox{traffic accidents}.
\noindent A standard technique to address this issue is anomaly detection \cite{Lis2019ICCV, Lis2020ARXIV, Biase2021CVPR, Chan2021ICCV, Tian2022ECCV, Grcic2022ECCV, Ackermann2023BMVC, Nayal2023ICCV, Delic2024ARXIV}, which differentiates between objects that fall into in-distribution (ID) classes a model knows from training and objects that do not (out-of-distribution (OOD) or outlier). Nevertheless, these anomaly detectors come with certain drawbacks. As illustrated in Fig.~\ref{fig:teaser}, compared with DNNs not designated for anomaly detection (left), many anomaly detectors \cite{Chan2021ICCV, Tian2022ECCV, Grcic2022ECCV, Nayal2023ICCV, Delic2024ARXIV} (middle) enhance the separability between ID and OOD by fine-tuning on OOD data. A procedure often referred to as outlier supervision guided by negative data. This approach necessitates extensive data collection and labeling as well as retraining of models and often sacrifices a small but non-negligible amount of performance on ID data. Moreover, these models can only reject OOD samples similar to the collected negative data and likely fail on other types of \mbox{OOD inputs.}\\\\
Seeking to avoid these drawbacks, we present a method called the \textbf{V}ision-\textbf{L}anguage Model for \textbf{A}nomaly \textbf{D}etection (VL4AD). It incorporates CLIP's \cite{Radford2021ICML} vision and text encoders into existing anomaly detectors. Vision-language models are typically exposed to a broader range of visual concepts during pre-training compared to the above mentioned vision-only models \cite{Radford2021ICML}. Previous work on image classification has shown that frozen CLIP models are as competitive as many sophisticated vision-only models in a zero-shot manner \cite{Ming2022NIPS}. We aim to leverage these advantageous generalization abilities of CLIP for improved pixel-level OOD-awareness without outlier supervision. Additionally, since vision-language models can handle textual input, we can utilize textual prompts to achieve data- and training-free outlier supervision, thereby increasing flexibility in real-world applications. Our contributions are as follows: (1)~we develop a method that applies FC-CLIP-type \cite{Yu2023NIPS} vision-language models to detect anomalous objects at the pixel level. (2)~Subsequently, we introduce a strategy that combines i)~max-logit prompt ensembling for a better alignment between the ID textual and visual embeddings with ii)~class merging to reduce the estimated uncertainty of edge pixels between ID class regions. \mbox{(3)~We} propose a new scoring function that enables data- and training-free outlier supervision via textual prompts.
We evaluate our models on RoadAnomaly19 (RA19) \cite{Lis2019ICCV}, FishyScapes Lost and Found (FS LaF) \cite{Blum2021IJCV}, and the Segment-Me-If-You-Can (SMIYC) dataset \cite{Chan2021NIPS}, achieving competitive performance.


\section{Related Work}
We begin by providing a brief overview of anomaly detection techniques in Section~\ref{ano_det}. Next, we summarize recent advancements in pixel-wise anomaly detection in Section~\ref{p_ano_det} and, finally, discuss vision-language pre-training in Section~\ref{vl_pretrain}.
\subsection{Anomaly Detection} \label{ano_det}
Anomaly detection aims to identify inputs that deviate from the training distribution, as these likely lead to erroneous predictions. In the literature, this task is also often referred to as out-of-distribution detection. We will use the terms out-of-distribution and anomaly interchangeably. A large body of research focuses on uncertainty-based methods like the softmax score \cite{Hendrycks2017ICLR}, Bayesian approximations \cite{Gal2016ICML, Lakshminarayanan2017NeurIPS} and feature-based methods, including the Mahalanobis distance \cite{Lee2018NeurIPS} and k-nearest neighbor approaches \cite{Sun2022ICML}. Recent work proposed using foundation models to detect OOD in large-scale setups. The authors of \cite{Ming2022NIPS}, for example, utilize normalized cosine similarities from CLIP models \cite{Radford2021ICML}, while \cite{Esmaeilpour2022AAAI, Wang2023ICCV} extend this approach by adding learnable text encoders. These methods primarily focus on image \mbox{classification.}

\subsection{Pixel-wise Anomaly Detection} \label{p_ano_det}
This task expands the previous anomaly detection task at the image level to the pixel level, typically focusing on semantic segmentation. As discussed in Chapter~\ref{sec:intro}, we aim to distinguish objects belonging to unknown semantic classes from those in known semantic categories. A related line of research, often referred to as industrial anomaly detection \cite{Zavrtanik2021PR, Yi2020ACCV, gu2024aaai, cai2024arxiv}, focuses instead on identifying irregular object fragments or incomplete objects of known semantic classes such as missing copper wires in cables or damaged surfaces of nuts. These types of anomalies present a different challenge, and as a result, methods from that domain are not directly applicable to our task. In the following, any mention of anomaly detection refers to the detection of unknown objects, as opposed to industrial anomaly detection.\\\\
Early methods for pixel-wise anomaly detection often utilized non-mask-based decoders with techniques such as image resynthesis \cite{Lis2019ICCV, Lis2020ARXIV, Biase2021CVPR}, auxiliary networks \cite{Besnier2021ICCV}, meta-classifiers \cite{Chan2021ICCV}, energy-based objectives \cite{Tian2022ECCV}, and mixtures of dense predictions \cite{Grcic2022ECCV}. Recent methods have adopted the popular mask-based decoder \cite{Cheng2022CVPR}. For instance, Maskomaly \cite{Ackermann2023BMVC} adds new post-processing steps, RbA (rejected by all) \cite{Nayal2023ICCV} implements an one-vs-all scoring function, and the recent UNO technique \cite{Delic2024ARXIV}, introduces an additional unknown class into their training objective. Regardless of the architecture, objective, or scoring functions, these models are trained based on a vision-only paradigm and they frequently fine-tune networks using use case-specific negative data. This sharpens the distinction between ID and OOD and thus enhances the model's discriminative capability; however, it also necessitates extensive data collection and labeling as well as model retraining.

\subsection{Vision-Language Pre-Training}\label{vl_pretrain}
Vision-only models are typically pre-trained on datasets such as ImageNet-1K \cite{DENG2009CVPR} or its extensive superset ImageNet-22K \cite{Russakovsky2015IJCV}.
Despite enabling enormous progress in computer vision over the last decade, 
these datasets contain only a few thousand semantic classes leading to limited diversity.
In contrast, CLIP \cite{Radford2021ICML} constructs a query list of 500K text items from Wikipedia to gather 400M image-text pairs. This approach allows CLIP to include far more rare visual concepts in its data compared to traditional computer-vision datasets like ImageNet. Similarly, LAION-2B \cite{Schuhmann2022NIPS} collects two billion text-image pairs. With such multi-modal data, CLIP-like models \cite{ZhaiCVPR2022, Jia2021ICML, Sun2023arxiv, Radford2021ICML} use contrastive learning \cite{Oord2018arxiv} to train the text and vision encoders. These models possess open-vocabulary capabilities, allowing them to make inferences on arbitrary semantic classes based solely on textual descriptions without retraining on these classes since the corresponding visual concepts are in many cases part of the pre-training corpus. Ming \textit{et al.} \cite{Ming2022NIPS} demonstrate that a frozen CLIP model, with simple normalized cosine similarity, is as strong as sophisticated vision-only anomaly detection methods \cite{Huang2021CVPR, Liu2020NeurIPS} in a zero-shot manner. This showcases how large-scale pre-training of models strongly improves sensitivity to OOD inputs.


\section{Methodology}
In this work, we introduce a pixel-wise anomaly detection method leveraging vision-language models. We begin by outlining how to integrate vision-language encoders into existing anomaly detectors in Section~\ref{vis_lang_enc}. Subsequently, we describe how to improve our model's effectiveness through max-logit prompt ensembling and class merging in Section~\ref{max_log_cls_meg}. Lastly, we introduce a scoring function for training- and data-free outlier supervision that exploits the open-vocabulary abilities of vision-language encoders in Section~\ref{method_ood_prompt}. Before presenting our method, let us define some basic notation for anomaly detection. We consider a semantic segmentation network trained for a $K$-class problem, accompanied by a pixel-wise anomaly detector $\mathcal{F}$. This detector is designed to process an input image $\mathbf{x} \in \mathbb{R}^{3\times H\times W}$ and produce an uncertainty score for each pixel, denoted as $\mathbf{u} \in \mathbb{R}^{H\times W}$, where $H$ and $W$ represent the image height and width, respectively. The primary function of the anomaly detector is to determine whether a pixel belongs to one of the known $K$ semantic classes (in-distribution, ID) or to an undefined class (out-of-distribution, OOD). This decision is made by thresholding the predicted uncertainty score \cite{Chan2021NIPS}. 

\subsection{Transitioning from Vision-Only to Vision-Language Encoders}\label{vis_lang_enc}
\begin{figure}[t]
\centering
\includegraphics[width=0.7\textwidth]{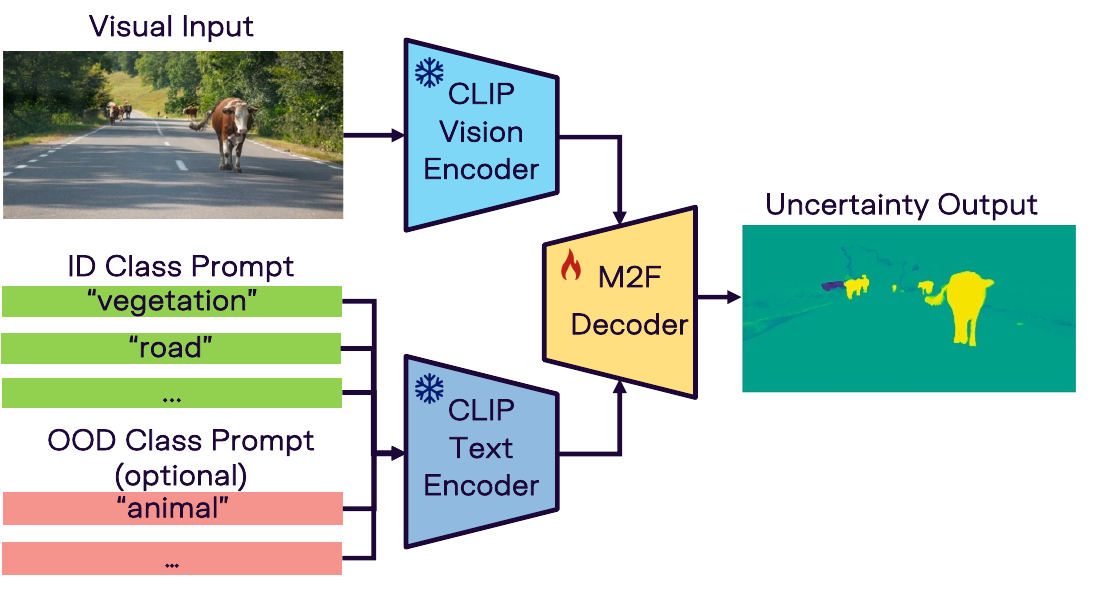}
\caption{\textbf{Our VL4AD approach uses the FC-CLIP architecture \cite{Yu2023NIPS}} It comprises frozen CLIP text and vision encoders paired with a Mask2Former (M2F) decoder. The model accepts visual inputs along with ID and optional OOD class prompts, providing pixel-wise uncertainty scores for anomaly detection.}
\label{fig:arch}
\end{figure}

Our method builds on the FC-CLIP model as detailed in \cite{Yu2023NIPS}. We repurpose this open-vocabulary semantic segmentation model for pixel-wise anomaly detection. An illustration of our architecture can be found in Fig.~\ref{fig:arch}. Unlike the typical vision encoder $\mathcal{E}_\text{vision, vis-only}$ - vision decoder $\mathcal{D}_\text{vis-only}$ architecture,  our vision encoder, $\mathcal{E}_\text{vision, vis-lang}$ is co-pre-trained with a text encoder, $\mathcal{E}_\text{text}$, which transforms textual prompts into embeddings that are subsequently fed to the decoder. An example of a textual prompt might be: "This is a photo of a \{\textit{class}\}.". \\\\ 
Consistent with previous work \cite{Ackermann2023BMVC, Nayal2023ICCV, Grcic2023CVPRW}, our decoder, $\mathcal{D}_\text{vis-lang}$, is based on Mask2Former \cite{Cheng2022CVPR}. It processes multi-scale visual and textual embeddings to produce two types of outputs: mask prediction scores $\mathbf{s} \in [0, 1]^{N\times H\times W}$ and mask classification scores $\mathbf{c} \in [0, 1]^{N\times K}$, where $N$ indicates the number of object queries. Object queries \cite{Zhu2021ICLR, Cheng2022CVPR} are learnable embeddings analogous to prior boxes in object detection networks \cite{He2017ICCV}. The mask prediction score identifies objects in a class-agnostic manner, whereas the mask classification score calculates the probability of masks belonging to specific semantic classes. Unlike \cite{Zhu2021ICLR}, which employs a linear transformation of visual embeddings, our approach calculates the mask classification score based on the cosine similarity between the processed visual embeddings $\mathbf{v}_i$, $i=1, \dots, N$, and ID class textual embeddings $\mathbf{t}_j$, $j=1, \dots, K$:
\begin{equation}\label{eq:cos}
    \mathbf{c}_{i} = \text{softmax}\Big(1/T
    \begingroup
    \setlength\arraycolsep{2pt}
    \begin{bmatrix}
    \text{cos}(\mathbf{v}_i, \mathbf{t}_1), & 
    \text{cos}(\mathbf{v}_i, \mathbf{t}_2), &
    \ldots, & 
    \text{cos}(\mathbf{v}_i, \mathbf{t}_{K})
    \end{bmatrix}
    \endgroup\Big)    
\end{equation}
where temperature $T$ is a trainable parameter that adjusts the sharpness of the softmax scaling. Following \cite{Grcic2023CVPRW}, the overall uncertainty score can be \mbox{expressed as}
\begin{equation}\label{eq: max_score}
    \mathbf{u}_{h,w} = -\max_{k}\sum_{i=1}^{N} \mathbf{s}_{i, h, w} \cdot \mathbf{c}_{i, k}\ \ .
\end{equation}
\begin{figure}
\centering
\includegraphics[width=1\textwidth]{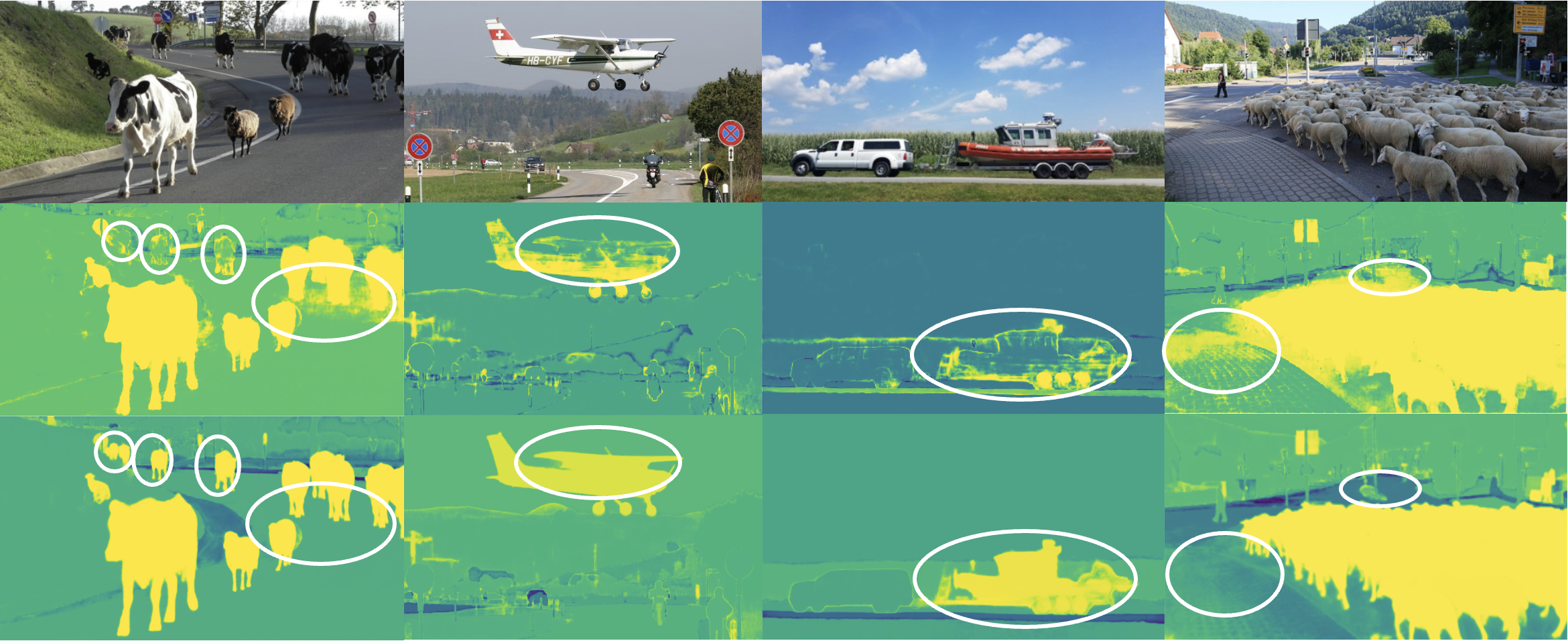}\\
\caption{\textbf{Comparison of VL4AD (bottom, ours) with RbA (middle) on four sample images (top)} Challenging OOD cases such as distant cows, airplanes and boat trailers are recognized with a notably cleaner and much more complete appearance (see white ellipses). While both methods successfully detect a flock of sheep as OOD (rightmost column), VL4AD produces far fewer false positives, such as misidentifying the road as an anomaly. Yellow indicates high ID class uncertainty (outliers), whereas blue signifies low ID class uncertainty (ID areas).}
\label{fig:visu}
\end{figure}\\
Architecturally, $\mathcal{E}_\text{vision, vis-only}$ and $\mathcal{E}_\text{vision, vis-lang}$, $\mathcal{D}_\text{vis-only}$ and $\mathcal{D}_\text{vis-lang}$ are quite similar. A key distinction is in the training approach: unlike $\mathcal{E}_\text{vision, vis-only}$, which is typically pre-trained on datasets like ImageNet-22K and then fine-tuned, $\mathcal{E}_\text{vision, vis-lang}$ remains frozen after vision-language pre-training and only the vision-language decoder $\mathcal{D}_\text{vis-lang}$ is fine-tuned. This way, we seek to transfer the competitive OOD detection performance of zero-shot CLIP \cite{Ming2022NIPS} from image-level to our pixel-level task.
\subsection{Max-Logit Prompt Ensembling and Class Merging} \label{max_log_cls_meg}
We improve the separability between ID and OOD categories by optimizing the ID class textual embeddings to better align with the corresponding ID visual embeddings. Instead of naively fine-tuning the text encoder—which can lead to catastrophic forgetting \cite{Ming2022NIPS} and may compromise the adaptability of textual prompts as highlighted in \cite{Minderer2022ECCV, KUO2023ICLR}, we use max-logit prompt ensembling \cite{Ghiasi2022ECCV, Xu2023CVPR, Yu2023NIPS}, which is originally devised to enhance generalization. We discover that max-logit prompt ensembling can significantly increase the model's sensitivity to OOD inputs by introducing concept lexical diversification and concretization in textual prompts. Lexical diversification includes synonyms and plural forms, while concretization involves decomposed concepts that align better with CLIP's pre-training \cite{Radford2021ICML}. We use, for example, the concepts \{\textit{vegetation, tree, trees, palm tree, bushes}\} to represent the class \textit{vegetation}. Max-logit ensembling allows us to consider all alternative concepts for a given class $k$ by replacing the term $\text{cos}(\mathbf{v}_i, \mathbf{t}_k)$ in Eq.~\ref{eq:cos} using the maximum cosine similarity between visual embeddings $\mathbf{v}_i$ and the textual embeddings of all $l$ alternative text embeddings $[\mathbf{t}_{k}^{1}, \ldots, \mathbf{t}_{k}^{l}]$ for the $k$-th class: 
\begin{equation}\label{eq:cos_max}
    \max\Big(
    \begingroup
    \setlength\arraycolsep{2pt}
    \begin{bmatrix}
    \text{cos}(\mathbf{v}_i, \mathbf{t}_{k}^{1}), & 
    \text{cos}(\mathbf{v}_i, \mathbf{t}_{k}^{2}), &
    \ldots, & 
    \text{cos}(\mathbf{v}_i, \mathbf{t}_{k}^{l})
    \end{bmatrix}
    \endgroup\Big).
\end{equation}
Additionally, solely relying on maximum pixel-wise scores along the $K$-class dimension can lead to suboptimal performance due to the high uncertainty of edge pixels between two ID classes, especially as the number of classes increases. To address this issue, we propose merging related ID classes into superclasses. This can be achieved by concatenating the textual prompts of individual semantic classes as different alternative concepts within superclasses during testing, without requiring retraining. The uncertainty of the superclasses can then be obtained using the max-logit fashion as described above.

\subsection{Data- and Training-Free Outlier Supervision via OOD Prompting}\label{method_ood_prompt}
\begin{figure}
\centering
\includesvg[width=1\textwidth]{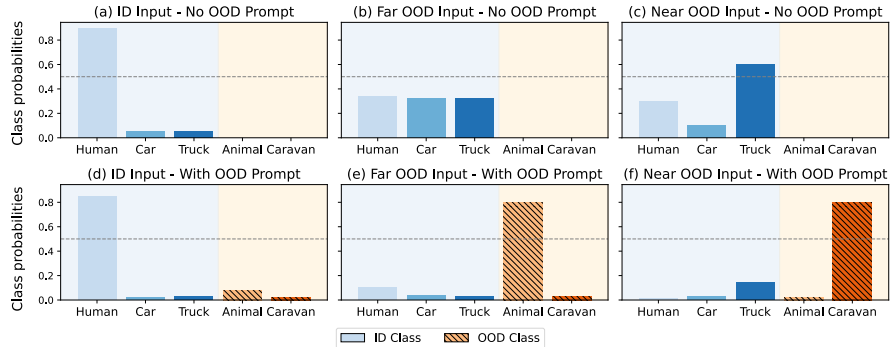}
\caption{\textbf{Comparison of VL4AD predictions without and with OOD prompting for an ID (left), far-OOD (middle) and near-OOD (right) input} We assume a simplified setup with three ID classes (\textit{human}, \textit{car}, \textit{truck}) and two OOD classes (\textit{animal}, \textit{caravan}). For an ID input (left), the model correctly predicts the class both without and with OOD prompts. For far-OOD (middle), the model also works well in both cases, however, using OOD prompts it puts significantly less weight on the (wrong) ID classes. For near-OOD, finally, the introduction of OOD classes is crucial as this way the erroneous classification of the input as ID (see panel c) can be avoided (panel f). Please note that an input is considered OOD when all its ID class probabilities (negative uncertainties) are below the decision threshold (dashed horizontal line). For further details, see Section~\ref{method_ood_prompt}.
}
\label{fig:out_sup}
\end{figure}

With vision-language pre-training, semantic OOD classes that are distinct from ID classes—often referred to as far-OOD—are typically well-detected. However, near-OOD cases, where the OOD class closely resembles an ID class, present more of a challenge. For instance, considering CityScapes \cite{Cordts2016CVPR} classes, the OOD class \textit{caravan} may appear visually similar to the ID class \textit{truck} in urban driving scenes. Leveraging the open-vocabulary capabilities of vision-language models, we introduce a new scoring function designed to better detect these near-OOD classes without requiring additional training or data preparation. To integrate $Q$ new OOD concepts at test time, the mask classification scores $\mathbf{c}_i$ from Eq.~\ref{eq:cos} need to be extended by the $Q$ additional terms $\text{cos}(\mathbf{v}_i, \mathbf{t}_{K+1}), \ldots, \text{cos}(\mathbf{v}_i, \mathbf{t}_{K+Q})$. Following Eq. \ref{eq: max_score}, i.e., by combining the first $K$ channels of $\mathbf{c} \in \left[0, 1\right]^{N\times (K+Q)}$ with the mask prediction score $\mathbf{s} \in \left[0, 1\right]^{N\times H\times W}$, we obtain the final uncertainty scores $\mathbf{u} \in \mathbb{R}^{H\times W}$.
With this integration, OOD objects from these $Q$ classes will (in most cases) be correctly assigned to their corresponding class. Without it, they would have been mistakenly assigned to an ID class similar to their actual near-OOD object class.
Conversely, if no OOD object is present in an input, the impact of the additional $Q$ classes remains negligible. For an illustration of these cases, see Fig.~\ref{fig:out_sup}.


\section{Experiments}
In this chapter, we first describe our experimental setup in Section \ref{exper_setup}. Next, we present the results on pixel-wise evaluation benchmarks in Section \ref{pixel_eval}. Ablation studies of our approach can be found in Section \ref{ablation}. We further investigate the impact of OOD prompting in Section \ref{impact_ood_prompting}, and, finally, conduct efficiency and trade-off analyses in Section~\ref{inference}.
\subsection{Experimental Setup} \label{exper_setup}
Following previous work, we evaluate our method on several pixel-wise anomaly datasets, namely, the SMIYC test set, RoadAnomaly19 (RA19), and the Fishy\-Scapes Lost and Found (FS LaF) validation set. All datasets treat object types from the widely used 19 CityScapes categories as ID and all other objects types as OOD. Notably, the SMIYC test set features two tracks: the Road Anomaly track (RA21) and the Road Obstacle track (RO21), the latter characterized by generally smaller anomalous objects. For pixel-level evaluation, we utilize average precision (AP) and false positive rate (FPR) at 95\% true positive rate. The SMIYC test suite additionally provides component-level metrics such as adjusted IoU over ground truth segments (sIoU gt) \cite{Rottmann2020IJCNN, Chan2021NIPS}, positive predicted value (PPV) \cite{Chan2021NIPS}, and mean F1, focusing on the completeness and integrity of detected OOD segments under various thresholds.\\\\
Unless otherwise stated, we use the hyperparameters values specified in \cite{Yu2023NIPS}. We adopt the ConvNext-L variant \cite{Liu2022CVPR} sourced from OpenCLIP \cite{Garbriel2021}, pre-trained on the LAION-2B dataset \cite{Schuhmann2022NIPS}. As stated in Section \ref{vis_lang_enc}, our encoder is frozen and our decoder is first trained on COCO panoptic \cite{Lin2014ECCV} as in \cite{Yu2023NIPS} for better initialization. Subsequent fine-tuning is done on the CityScapes dataset for 10K iterations at a learning rate of \(2.5 \times 10^{-5}\), with reductions by a factor of ten at 85\% and 95\% of the duration. For max-logit ensembling, we mainly adopt the conceptual dictionary as outlined in \cite{Ghiasi2022ECCV, Xu2023CVPR, Yu2023NIPS} with some modifications. For OOD prompting, we use the names of the OOD classes from the RA21 dataset for evaluations on RA21 and RO21, and those from RA19 for assessments on RA19 and FS LaF. These class names are standard metadata associated with each dataset. More details on the training hyperparameters, prompt descriptions, and class-merging strategies can be found in Appendix \ref{training}, \ref{ensembling}, and \ref{merging}, respectively. To ensure a fair comparison, we also evaluate VL4AD with OOD prompting against other methods that utilize outlier exposure. However, as previously mentioned, these techniques build on OOD image datasets, whereas our approach leverages only the textual names of OOD classes without requiring any visual information. We train and evaluate our models using three random seeds and report the mean performance with two standard deviations.

\subsection{Pixel-wise Evaluation}\label{pixel_eval}
Tab. \ref{tab:smiyc_merged} presents the evaluation results on RoadAnomaly19 (RA19), FishyScapes Lost and Found (FS LaF), and both the anomaly (RA21) and obstacle (RO21) tracks of SMIYC. On RA19, VL4AD surpasses competing methods, including recent works RbA \cite{Nayal2023ICCV} and UNO \cite{Delic2024ARXIV} with and without outlier supervision. Utilizing VL4AD with OOD prompting yields an additional gain of 3.3\% in AP and a reduction of 0.6 in FPR, achieved without further training or data preparation. On FS LaF, VL4AD achieves a superior FPR compared with UNO \cite{Delic2024ARXIV} without outlier supervision; however, UNO records a marginally higher AP. Notably, the performance boost from including OOD prompts observed on RA19 does not extend to FS LaF, attributable to the prompts being specifically tailored for RA19. Consequently, VL4AD ranks as a close second among methods that leverage outlier supervision. Please note that the other models do not use class merging. For test results of VL4AD using the standard 19 classes, see the ablation study in Section~\ref{ablation}. Moreover, we compare VL4AD with the original FC-CLIP \cite{Yu2023NIPS} model on FS LaF and RA19 and find it to be significantly weaker then VL4AD, emphasizing the importance of our modifications.\\\\
Analyzing the anomaly track (RA21) of SMIYC \cite{Chan2021NIPS}, VL4AD is competitive at the component level (sIoU gt, PPV and mean F1) across all models, with and without outlier supervision. However, at the pixel level, VL4AD without OOD prompting suffers from an exceptionally high FPR. Upon manual inspection, we identified an unusual pattern where caravans are mistakenly classified as trucks due to their visual resemblance, assigning them low uncertainty values. This misclassification affects approximately 10\% of the predictions, significantly elevating the FPR. To address this, introducing "caravan" as one of the OOD prompts substantially alleviates the issue, reducing the FPR by a factor of 25 and improving the AP by $10.4\%$. This improvement positions VL4AD as the second-best model in settings with outlier supervision, underscoring the effectiveness and flexibility of OOD prompting. For the obstacle track (RO21), VL4AD does not perform as well, potentially due to the small size of the anomalies which are difficult for the encoder to represent, as it was originally trained on object-centered image-text pairs. We delve further into this issue in Appendix \ref{obstacle}.
\begin{table}[t!]
    \caption{\textbf{Results on RA19, FS LaF and the SMIYC test track} Both pixel-level (AP, FPR) and component-level metrics (sIoU, PPV, mean F1) are reported on the SMIYC anomaly and the obstacle track. On RA19 and FS LaF, we report pixel-level metrics (AP and FPR). Results are shown with the best in each group in bold. The top two groups correspond to vision-only models without and with outlier supervision ("out. sup."), respectively, while the bottom group refers to vision-language models.} 
    \adjustbox{max width=0.99\linewidth}{%
    \begin{tabular}{l c c c c c c c c c c c c c c c } 
    \toprule
     \multirow{2}{*}{method} & \multirow{2}{*}{\raisebox{-0.5ex}{\shortstack{out. \\[0.4em] sup.}}} & \multicolumn{2}{c}{RA19}& \multicolumn{2}{c}{FS LaF}& \multicolumn{5}{c}{SMIYC RA21}  & \multicolumn{5}{c}{SMIYC RO21} \\       \cmidrule(r){3-4} \cmidrule(lr){5-6} \cmidrule(lr){7-11}\cmidrule(l){12-16}
           &  & AP $\uparrow$  & FPR $\downarrow$ & AP $\uparrow$  &FPR $\downarrow$ &  AP $\uparrow$ & FPR $\downarrow$ & sIoU gt $\uparrow$ & PPV $\uparrow$ & mF1 $\uparrow$ & AP $\uparrow$  & FPR $\downarrow$ & sIoU gt $\uparrow$ & PPV $\uparrow$ & mF1 $\uparrow$  \\
    \midrule
     Max. Softmax \cite{Hendrycks2017ICLR}& \xmark & 20.6 & 68.4  & 6.0  &45.6 & 28.0&72.1&15.5&15.3&5.4&15.7  & 16.6 & 19.7&15.9 &6.3 \\
      Image Resyn. \cite{Lis2019ICCV}                & \xmark & -& - & - &-& 52.3 & 25.9 & 39.7 & 11.0 & 12.5 & 37.7 & 4.7 & 16.6 & 20.5 & 8.4  \\ 
      ObsNet \cite{Besnier2021ICCV}                  & \xmark & - & -& -&-& 75.4 & 26.7 & 44.2 & \textbf{52.6} & 45.1 & - & - & - & - & -   \\
      Maskomaly \cite{Ackermann2023BMVC} & \xmark & 80.8 & 12.0 & - &-& 93.4& 6.9& 55.4 & 51.5 & 49.9 & - & - & - & - & -  \\
      RbA \cite{Nayal2023ICCV}                       & \xmark & 78.5  & 11.8  & 61.0 &10.6 & 86.1& 15.9& 56.3 & 41.4 & 42.0 &87.8  & 3.3 & 47.4& 56.2& 50.4\\
                UNO \cite{Delic2024ARXIV}  & \xmark & \textbf{82.4}  & \textbf{9.2}  & \textbf{74.5} &\textbf{6.9}& \textbf{96.1}&\textbf{2.3}&\textbf{68.0}&51.9&\textbf{58.9}& \textbf{89.0}&\textbf{0.6}&\textbf{66.9}&\textbf{74.9}&\textbf{76.3}\\
     \midrule
      Max. Entropy \cite{Chan2021ICCV}                 & \cmark & -     & -     & 41.3 &37.7 & 85.5 & 15.0 & 49.2 & 39.5 & 28.7 & 85.1 & 0.8 & 47.9 & 62.6 & 48.5  \\ 
      DenseHybrid \cite{Grcic2022ECCV}               & \cmark & - & - & 63.8 &6.1& 78.0 & 9.8 & 54.2 & 24.1 & 31.1 & 87.1 & \textbf{0.2}& 45.7 & 50.1 & 50.7   \\ 
      PEBAL \cite{Tian2022ECCV}                      & \cmark & 44.4 & 38.0  & 64.4 &6.6  & 49.1 & 40.8 & 38.9 & 27.2 & 14.5 & 5.0 & 12.7 & 29.9 & 7.6 & 5.5  \\
      SynBoost \cite{Biase2021CVPR}                  & \cmark & 38.2 & 64.8  & 60.6 &31.0 & 56.4 & 61.9 & 34.7 & 17.8 & 10.0 & 71.3 & 3.2  & 44.3 & 41.8 & 37.6  \\
      RbA \cite{Nayal2023ICCV}                       & \cmark & 85.4 & \textbf{6.9}& 70.8 &6.3 & 90.9 &11.6 &55.7 &52.1& 46.8& 91.8&0.5& 58.4&58.8& 60.9\\ 
      UNO \cite{Delic2024ARXIV}        & \cmark & \textbf{88.5}& 7.4  & \textbf{81.8} &\textbf{1.3} & \textbf{96.3} & \textbf{2.0} & \textbf{68.5} & \textbf{55.8}  & \textbf{62.6}& \textbf{93.2}&\textbf{0.2}&\textbf{71.1}&\textbf{72.2}&\textbf{77.7}\\
    \midrule
    FC-CLIP \cite{Yu2023NIPS} & \xmark &                 75.9   &  74.2 &   6.7 &    89.0& -	& - & -& - & -& -	& -& -&- & -   \\
          VL4AD (ours)                                   & \xmark &                88.9 & 6.4 & 73.7 &\textbf{3.9}& 82.5	& 82.4 & 67.0& 51.9 & 61.1& 76.4	& 0.6& 42.5&19.0 & 23.6   \\
                     VL4AD w. prompt (ours)                                & \xmark & \textbf{92.2}  & \textbf{5.8} & \textbf{73.8}&4.6& \textbf{92.9}& \textbf{3.3}& \textbf{71.6}	& \textbf{53.7} & \textbf{65.4} & 78.7&	0.6& 36.4	&22.4& 24.4  \\
     \bottomrule
    \end{tabular}
    }
    \label{tab:smiyc_merged}
\end{table} 

\subsection{Ablation Study}\label{ablation}
We conduct an ablation study to validate the effectiveness of our model design, including pre-training, max-logit prompt ensembling, and class merging. An additional analysis of OOD prompting will be provided in the following section. For pre-training, we utilize the vision-only ConvNext-L model \cite{Liu2022CVPR} pre-trained on ImageNet-22K, along with the standard Mask2Former decoder. We use the same training protocol (COCO-CityScapes) as before for the vision-language variant. We also experimented with training the model on CityScapes for 90K iterations from scratch, as in \cite{Liu2022CVPR} and \cite{Nayal2023ICCV}, but did not observe any significant benefit over the current setup. We conduct two sets of experiments with unfrozen and frozen encoders, respectively. As for max-logit ensembling, we compare our approach with a baseline that uses only one concept for each semantic class.
The results of all these ablations can be found in Tab.~\ref{tab:abl} that shows pixel-level metrics on RA19 and FS LaF, as well as mIoU as a generalization score on CityScapes. For vision-only models, the unfrozen variant provides the best CityScapes classification performance across all setups; however, both variants exhibit low AP on both OOD datasets, indicating a lack of OOD awareness. With vision-language pre-training, our model surpasses vision-only models in every metric except for FPR on RA19. This validates our hypothesis that vision-language models are often more OOD-aware than vision-only models. Class merging reduces the FPR on RA19 by almost a factor of three, to 15.3\%, and the FPR on FS LaF by a factor of two, to 4.2\%. This demonstrates its effectiveness in addressing high uncertainty between ID classes. The max-logit prompt ensembling slightly improves mIoU on CityScapes by 0.8\%. On anomaly detection in RA19, it significantly improves AP by 13.8\%, which can be attributed to the better alignment between the ID visual embeddings and the text embeddings. Combining all three methods, we achieve the best performance without any outlier supervision, validating the design of our models.

\subsection{Impact of OOD Prompting}\label{impact_ood_prompting}
Previously, we observed the benefits of OOD prompts on the RA19 and SMIYC RA21 datasets in Tab.~\ref{tab:smiyc_merged}. This can be attributed to the use of multiple OOD concepts. In this section, we investigate the effects of individual prompts. For a given type of OOD concept, we split the RA19 dataset into two subsets, one containing all frames that show this concept (known unknowns) and a complementary one that shows other OOD concepts but not the selected one (unknown unknowns). We measure the performance of VL4AD with OOD prompting for the whole dataset and both subsets. Additionally, we include the performance metrics on all subsets without OOD prompting for comparison. As shown in Tab.~\ref{tab:ra_val}, AP and FPR on the known unknowns subsets are generally better than the counterpart without OOD prompting. This indicates that our model retains open-vocabulary capabilities and can reject these OOD objects with our scoring function introduced in Section~\ref{method_ood_prompt}. Regarding the unknown-unknowns subsets, we observe fluctuations around 1\% in AP and FPR when applying different prompts, but these are not statistically significant. We conclude that OOD prompts positively affect OOD detection on the known unknowns while having a negligible impact on the unknown unknowns. These results are expectable since the OOD prompts only trigger higher ID class uncertainty when corresponding OOD objects show up (see Section~\ref{method_ood_prompt}). We also experiment with adopting the OOD class names of RbA's outlier supervision dataset as OOD prompts (see last row of Tab.~\ref{tab:ra_val}). Some of these classes overlap with instances in RA19, such as \textit{cow, zebra,} and \textit{bear}, which explains why the RbA prompts perform similarly to the prompts used in the previously described experiments. For details on the RbA outlier classes, please refer to Section~\ref{max-logit_concept}. These results again underline the flexibility and effectiveness of OOD prompting.

\begin{table}[t]
\caption{\textbf{Ablation study} We validate the effectiveness of our model design, including vision-language pre-training, max-logit prompt ensembling, and class merging on two anomaly detection datasets: RA19 and FS LaF. To assess generalization, we also report the mean Intersection over Union (mIoU) on the CityScapes dataset. Note that the mIoU on CityScapes cannot be calculated when class merging is applied.}
\footnotesize
\centering
\adjustbox{max width=\textwidth}{%
    \begin{tabular}{c c c c c c c c c c cc} 
     \toprule
     \multirow{2}{*}{\shortstack{\rule{0pt}{8pt} vis-lang \\[0.1em]pre-training}} & \multirow{2}{*}{\shortstack{\rule{0pt}{8pt} max-logit\\[0.1em]ensembling}} & \multirow{2}{*}{\shortstack{\rule{0pt}{8pt} class\\[0.32em]merging}} & \multicolumn{4}{c}{RA19} & \multicolumn{4}{c}{FS LaF} & CityScapes\\
      \cmidrule(r){4-7} \cmidrule(l){8-11} \cmidrule(l){12-12} 
          & & & \multicolumn{2}{c}{AP $\uparrow$}  & \multicolumn{2}{c}{FPR $\downarrow$ }&\multicolumn{2}{c}{ AP $\uparrow$ } & \multicolumn{2}{c}{FPR $\downarrow$}   & mIoU $\uparrow$ \\
     \midrule
       \xmark\ (unfrozen)     & \xmark     &\xmark      & 55.8 & \tiny{$\pm 6.0$}   &   31.8  & \tiny{$\pm 10.5$}  &    61.3  &  \tiny{$\pm 7.9$}  &    10.7 &  \tiny{$\pm 2.4$}   &\textbf{80.9} \\
       \xmark\ (frozen)     & \xmark     & \xmark      & 62.9  &  \tiny{$\pm 5.7$}  &     27.4  &  \tiny{$\pm 2.1$}  &   41.8  &   \tiny{$\pm 13.7$}    &   9.3  &  \tiny{$\pm 2.5$}   &79.7  \\
        \cmark  & \xmark     & \xmark      & 72.3 & \tiny{$\pm 2.8$} &43.8 & \tiny{$\pm 3.8$} & 70.6 & \tiny{$\pm 2.7$} & 9.1 & \tiny{$\pm 1.1$} &77.1  \\
         \cmark  & \xmark     & \cmark  &75.9  & \tiny{$\pm 2.6$}  &15.3  & \tiny{$\pm 1.5$} &\textbf{74.5} & \tiny{$\pm 3.0$} & 4.2  & \tiny{$\pm 0.7$}  &- \\
          \cmark  & \cmark  & \xmark      &  86.1   & \tiny{$\pm 0.3$}   &  14.4 & \tiny{$\pm 2.7$} &   70.4  &  \tiny{$\pm 2.8$} &    9.8  &  \tiny{$\pm 0.6$}  &77.9  \\
            \cmark  & \cmark  & \cmark  & \textbf{88.9}   &  \tiny{$\pm 0.5$}  &  \textbf{6.4}  &  \tiny{$\pm 2.1$}  &  73.7  & \tiny{$\pm 2.3$}  & \textbf{3.9}  &  \tiny{$\pm 0.2$} &-\\
        \bottomrule
    \end{tabular}}
    \vspace{-.05in}
    \label{tab:abl}
\end{table}

\begin{table*}[t]
\footnotesize
    \centering
    \caption{\textbf{Impact of different OOD prompts for VL4AD on RA19} We show results on the whole dataset, the subset containing the OOD prompt concept (known unknowns) and the complementary subset without this concept (unknown unknowns). Performance scores without OOD prompting are provided in parentheses.}
    \adjustbox{max width=\textwidth}{
    \begin{tabular}{l c c c c c c} 
     \toprule
         \multirow{2}{*}{\shortstack{\rule{0pt}{10pt}OOD\\[0.3em]prompt}}  & \multicolumn{2}{c}{RA19} & \multicolumn{2}{c}{RA19 - known unknowns} & \multicolumn{2}{c}{RA19 - unknown unknowns}\\
        \cmidrule(r){2-3} \cmidrule(l){4-5} \cmidrule(l){6-7}
         & AP $\uparrow$ & FPR $\downarrow$   & AP $\uparrow$ & FPR $\downarrow$ & AP $\uparrow$ & FPR $\downarrow$\\
        \midrule
         - & 88.9  &6.4  &-&-&-&-\\
         \midrule
 animal&91.7 & 5.7  &  97.1 (94.8)& 3.1 (3.6)&71.5 (72.7)&7.6 (7.9)\\ 
        cone                    & 89.0 & 6.2 & 69.1 (62.6)& 5.6 (9.1)& 90.7 (90.2)& 5.9 (5.2)\\ 
        boulder     & 89.4 & 5.8 & 89.0 (89.0)& 12.5 (14.7)& 90.3 (89.9)& 4.6 (4.0)\\
        cardboard& 89.6 & 5.4 & 89.0 (82.7)& 3.0 (10.8)& 89.9 (89.6)& 5.2 (5.2)\\
         tire&89.8 & 5.2 & 18.0\hspace{2mm} (7.4)& 3.8 (67.2)& 91.2 (90.7)& 5.2 (5.2)\\
         \midrule
         all above&92.2  &    5.8&-&-&-&-\\
         RbA prompt&91.8&   4.5&-&-&-&-\\
         \bottomrule   
        \end{tabular}}
        \vspace{-.05in}
        \label{tab:ra_val}
\end{table*}

\subsection{Efficiency and Trade-Off Analysis}\label{inference}
\begin{figure}[t]
    \centering
        \centering
        \adjustbox{width=0.37\textwidth}{
\begin{tikzpicture}

\definecolor{darkgray176}{RGB}{176,176,176}
\definecolor{darkorange25512714}{RGB}{255,127,14}
\definecolor{lightgray204}{RGB}{204,204,204}
\definecolor{steelblue31119180}{RGB}{31,119,180}

\begin{axis}[
legend cell align={left},
font=\LARGE,
legend style={fill opacity=0.8, draw opacity=1, text opacity=1, draw=lightgray204},
tick align=outside,
tick pos=left,
x grid style={darkgray176},
xlabel={OOD Recall on RA19},
xmin=0, xmax=1,
xtick style={color=black},
y grid style={darkgray176},
ylabel={ID Pixel Retention Rate},
ymin=0.95, ymax=1.02,
ytick style={color=black}
]
\addplot [draw=steelblue31119180, fill=steelblue31119180, forget plot, mark=*, only marks]
table{%
x  y
0 1
1.83607082312707e-07 0.999949585531203
0.313229459462586 0.999496641555719
0.549447315141176 0.998471259830836
0.718227023914639 0.997129417747214
0.772985265715252 0.995494498693799
0.828192426832119 0.993244750161193
0.852866097374547 0.990204600989239
0.875975802789836 0.985831091569191
0.877253157261486 0.984663901362189
0.88306175091753 0.979709968530416
0.938479876571975 0.969038220776811
0.984526880352261 0.954131189823807
0.988259428728596 0.940505285711857
0.98996348605954 0.920809422196939
0.991027856315707 0.886204975960959
0.991444460785475 0.850011437446601
0.991788173243564 0.813081402331464
0.992033105091369 0.775848937109053
0.992123256168785 0.738494887642336
0.992217262994929 0.700939340602543
0.992297866504064 0.66318290121193
0.992297866504064 0.625314600390789
0.99236102734038 0.587408875024329
0.99236102734038 0.549475078250031
0.99236102734038 0.511515887222204
0.992403807790559 0.47353740650697
0.992403807790559 0.435574261360394
0.992403807790559 0.397624532519104
0.99242694228293 0.359743486044369
0.99242694228293 0.32197368705398
0.992469722733109 0.284119902843093
0.992469722733109 0.246296242343266
0.992520765501992 0.208565176486861
0.992747336641566 0.171310397646737
0.993777739587505 0.135011731481021
0.994643997801856 0.0993699571961411
0.99468163725373 0.0982187623045842
0.996262861446607 0.0648775196069138
0.996551491780003 0.0596086879988741
0.997699036044457 0.0450241957989573
0.998685556897723 0.0366528106065264
0.999023210322096 0.0317989488214125
0.999223158434735 0.0295472352248287
0.99939978844792 0.0244593541668493
0.999569074177812 0.0196073516681298
0.999828143770955 0.0156226315737893
0.999876616040686 0.00931923194844112
0.999913521064231 0.00521914994889487
1 1.09049055389329e-09
};
\addplot [draw=darkorange25512714, fill=darkorange25512714, forget plot, mark=*, only marks]
table{%
x  y
0 1
0.116929803156683 0.99994959970758
0.535730030021594 0.999533031225502
0.722315586423578 0.998464439902912
0.757184958246831 0.997079137408755
0.829053360441083 0.995378232771924
0.853195855694381 0.993070668611132
0.935691068598727 0.989893473129308
0.95431947595601 0.98527365133638
0.955082546990102 0.984593307365693
0.962060534153396 0.978549543726811
0.98591293021664 0.966750586421382
0.989239706941064 0.953915154921157
0.992112606958011 0.932609732801145
0.992934615865525 0.920266085278462
0.994226475296677 0.884705635417128
0.994555866402346 0.848591558768451
0.994756732550396 0.811697724668232
0.994859368909409 0.774461790595369
0.994915552676596 0.736998415088662
0.994975592192513 0.699389798235573
0.995018739856856 0.661721672222467
0.99504921863252 0.623940529949337
0.995152222205698 0.586202324651275
0.995189494443407 0.548473401608809
0.995199592832934 0.510722966459186
0.995202163332087 0.472981581290669
0.995207671544556 0.435250831676525
0.99522621585987 0.397532714304957
0.995229520787351 0.359785332528884
0.995235947035232 0.322000773748849
0.995248248709747 0.284178757708777
0.995265507775485 0.246553591571042
0.995270832380872 0.208848227486502
0.995349232605019 0.171567701073909
0.995515580621594 0.135428269429364
0.996505039188178 0.0994656662805847
0.996560121312871 0.0951874450156261
0.997405448319839 0.064938175962993
0.997723272179322 0.0582443087469745
0.99872099306461 0.0441248126241433
0.999081046553025 0.0358788251214856
0.999248312605012 0.0317920122109992
0.999335342362028 0.0288057060101435
0.999559526609532 0.0237313437635607
0.999808314206066 0.0188171527695337
0.999864865187418 0.015012431227
0.999890386571859 0.00859016049784357
0.999919212883782 0.00487586679400092
1 1.09049055389329e-09
};
\addplot [semithick, steelblue31119180]
table {%
0 1
1.83607082312707e-07 0.999949585531203
0.313229459462586 0.999496641555719
0.549447315141176 0.998471259830836
0.718227023914639 0.997129417747214
0.772985265715252 0.995494498693799
0.828192426832119 0.993244750161193
0.852866097374547 0.990204600989239
0.875975802789836 0.985831091569191
0.877253157261486 0.984663901362189
0.88306175091753 0.979709968530416
0.938479876571975 0.969038220776811
0.984526880352261 0.954131189823807
0.988259428728596 0.940505285711857
0.98996348605954 0.920809422196939
0.991027856315707 0.886204975960959
0.991444460785475 0.850011437446601
0.991788173243564 0.813081402331464
0.992033105091369 0.775848937109053
0.992123256168785 0.738494887642336
0.992217262994929 0.700939340602543
0.992297866504064 0.66318290121193
0.992297866504064 0.625314600390789
0.99236102734038 0.587408875024329
0.99236102734038 0.549475078250031
0.99236102734038 0.511515887222204
0.992403807790559 0.47353740650697
0.992403807790559 0.435574261360394
0.992403807790559 0.397624532519104
0.99242694228293 0.359743486044369
0.99242694228293 0.32197368705398
0.992469722733109 0.284119902843093
0.992469722733109 0.246296242343266
0.992520765501992 0.208565176486861
0.992747336641566 0.171310397646737
0.993777739587505 0.135011731481021
0.994643997801856 0.0993699571961411
0.99468163725373 0.0982187623045842
0.996262861446607 0.0648775196069138
0.996551491780003 0.0596086879988741
0.997699036044457 0.0450241957989573
0.998685556897723 0.0366528106065264
0.999023210322096 0.0317989488214125
0.999223158434735 0.0295472352248287
0.99939978844792 0.0244593541668493
0.999569074177812 0.0196073516681298
0.999828143770955 0.0156226315737893
0.999876616040686 0.00931923194844112
0.999913521064231 0.00521914994889487
1 1.09049055389329e-09
};
\addlegendentry{VL4AD}
\addplot [semithick, darkorange25512714]
table {%
0 1
0.116929803156683 0.99994959970758
0.535730030021594 0.999533031225502
0.722315586423578 0.998464439902912
0.757184958246831 0.997079137408755
0.829053360441083 0.995378232771924
0.853195855694381 0.993070668611132
0.935691068598727 0.989893473129308
0.95431947595601 0.98527365133638
0.955082546990102 0.984593307365693
0.962060534153396 0.978549543726811
0.98591293021664 0.966750586421382
0.989239706941064 0.953915154921157
0.992112606958011 0.932609732801145
0.992934615865525 0.920266085278462
0.994226475296677 0.884705635417128
0.994555866402346 0.848591558768451
0.994756732550396 0.811697724668232
0.994859368909409 0.774461790595369
0.994915552676596 0.736998415088662
0.994975592192513 0.699389798235573
0.995018739856856 0.661721672222467
0.99504921863252 0.623940529949337
0.995152222205698 0.586202324651275
0.995189494443407 0.548473401608809
0.995199592832934 0.510722966459186
0.995202163332087 0.472981581290669
0.995207671544556 0.435250831676525
0.99522621585987 0.397532714304957
0.995229520787351 0.359785332528884
0.995235947035232 0.322000773748849
0.995248248709747 0.284178757708777
0.995265507775485 0.246553591571042
0.995270832380872 0.208848227486502
0.995349232605019 0.171567701073909
0.995515580621594 0.135428269429364
0.996505039188178 0.0994656662805847
0.996560121312871 0.0951874450156261
0.997405448319839 0.064938175962993
0.997723272179322 0.0582443087469745
0.99872099306461 0.0441248126241433
0.999081046553025 0.0358788251214856
0.999248312605012 0.0317920122109992
0.999335342362028 0.0288057060101435
0.999559526609532 0.0237313437635607
0.999808314206066 0.0188171527695337
0.999864865187418 0.015012431227
0.999890386571859 0.00859016049784357
0.999919212883782 0.00487586679400092
1 1.09049055389329e-09
};
\addlegendentry{VL4AD w. prompt}
\end{axis}

\end{tikzpicture}}
        \adjustbox{width=0.37\textwidth}{
\begin{tikzpicture}

\definecolor{darkgray176}{RGB}{176,176,176}
\definecolor{darkorange25512714}{RGB}{255,127,14}
\definecolor{lightgray204}{RGB}{204,204,204}
\definecolor{steelblue31119180}{RGB}{31,119,180}

\begin{axis}[
font=\LARGE,
legend cell align={left},
legend style={fill opacity=0.8, draw opacity=1, text opacity=1, draw=lightgray204},
tick align=outside,
tick pos=left,
x grid style={darkgray176},
xlabel={OOD Recall on FS LaF},
xmin=0, xmax=1,
xtick style={color=black},
y grid style={darkgray176},
ylabel={ID Pixel Retention Rate},
ymin=0.95, ymax=1.02,
ytick style={color=black}
]
\addplot [draw=steelblue31119180, fill=steelblue31119180, forget plot, mark=*, only marks]
table{%
x  y
0 1
0.00109700542640852 0.999949585531203
0.345089853955919 0.999496641555719
0.580336805788169 0.998471259830836
0.660543813224359 0.997129417747214
0.700398187847525 0.995494498693799
0.746445199973203 0.993244750161193
0.819423276612849 0.990204600989239
0.846011003550613 0.985831091569191
0.852758424331748 0.984663901362189
0.870978344610437 0.979709968530416
0.896720288738527 0.969038220776811
0.929904702887385 0.954131189823807
0.942825919474777 0.940505285711857
0.954382159844577 0.920809422196939
0.962410815971059 0.886204975960959
0.966545521538152 0.850011437446601
0.973801668118175 0.813081402331464
0.974963572720573 0.775848937109053
0.975947527969451 0.738494887642336
0.976307613720105 0.700939340602543
0.976558836336839 0.66318290121193
0.976558836336839 0.625314600390789
0.976558836336839 0.587408875024329
0.976766094995645 0.549475078250031
0.976766094995645 0.511515887222204
0.976766094995645 0.47353740650697
0.97691473504388 0.435574261360394
0.97691473504388 0.397624532519104
0.97691473504388 0.359743486044369
0.977105245528237 0.32197368705398
0.977105245528237 0.284119902843093
0.977433928451799 0.246296242343266
0.977812855898707 0.208565176486861
0.979301349902861 0.171310397646737
0.983212048636699 0.135011731481021
0.988253249145843 0.0993699571961411
0.988387234541435 0.0982187623045842
0.992364925973069 0.0648775196069138
0.993263046827896 0.0596086879988741
0.995729215515509 0.0450241957989573
0.996907868292356 0.0366528106065264
0.99751080257252 0.0317989488214125
0.997799708581765 0.0295472352248287
0.998281218597173 0.0244593541668493
0.998628743216989 0.0196073516681298
0.99888205935553 0.0156226315737893
0.999170965364775 0.00931923194844112
0.99956245394252 0.00521914994889487
1 1.09049055389329e-09
};
\addplot [draw=darkorange25512714, fill=darkorange25512714, forget plot, mark=*, only marks]
table{%
x  y
0 1
0.00110747303543914 0.99994959970758
0.347960072352114 0.999533031225502
0.586979550478998 0.998464439902912
0.662610119247002 0.997079137408755
0.701953674549474 0.995378232771924
0.747914852281101 0.993070668611132
0.821292791585717 0.989893473129308
0.849812839150533 0.98527365133638
0.852932186641656 0.984593307365693
0.877409643598848 0.978549543726811
0.903417464996315 0.966750586421382
0.929900515843773 0.953915154921157
0.948474241307697 0.932609732801145
0.952879011187781 0.920266085278462
0.961424767200375 0.884705635417128
0.965643213639713 0.848591558768451
0.973441582367522 0.811697724668232
0.976596519729349 0.774461790595369
0.977450676626248 0.736998415088662
0.977942654250687 0.699389798235573
0.978344610437462 0.661721672222467
0.97875075366785 0.623940529949337
0.978914048368728 0.586202324651275
0.979236450726871 0.548473401608809
0.979391371340524 0.510722966459186
0.979531637301534 0.472981581290669
0.979682370871575 0.435250831676525
0.979799608092718 0.397532714304957
0.979799608092718 0.359785332528884
0.979921032357473 0.322000773748849
0.98004664366584 0.284178757708777
0.980161787365177 0.246553591571042
0.980536527768473 0.208848227486502
0.982054331077913 0.171567701073909
0.985644720975414 0.135428269429364
0.988761974944731 0.0994656662805847
0.989224643263884 0.0951874450156261
0.992946925035171 0.064938175962993
0.993891103369733 0.0582443087469745
0.996005560393917 0.0441248126241433
0.997064882427815 0.0358788251214856
0.997533831312387 0.0317920122109992
0.997881355932203 0.0288057060101435
0.998337743685938 0.0237313437635607
0.998637117304214 0.0188171527695337
0.99888205935553 0.015012431227
0.999196087626449 0.00859016049784357
0.999587576204194 0.00487586679400092
1 1.09049055389329e-09
};
\addplot [semithick, steelblue31119180]
table {%
0 1
0.00109700542640852 0.999949585531203
0.345089853955919 0.999496641555719
0.580336805788169 0.998471259830836
0.660543813224359 0.997129417747214
0.700398187847525 0.995494498693799
0.746445199973203 0.993244750161193
0.819423276612849 0.990204600989239
0.846011003550613 0.985831091569191
0.852758424331748 0.984663901362189
0.870978344610437 0.979709968530416
0.896720288738527 0.969038220776811
0.929904702887385 0.954131189823807
0.942825919474777 0.940505285711857
0.954382159844577 0.920809422196939
0.962410815971059 0.886204975960959
0.966545521538152 0.850011437446601
0.973801668118175 0.813081402331464
0.974963572720573 0.775848937109053
0.975947527969451 0.738494887642336
0.976307613720105 0.700939340602543
0.976558836336839 0.66318290121193
0.976558836336839 0.625314600390789
0.976558836336839 0.587408875024329
0.976766094995645 0.549475078250031
0.976766094995645 0.511515887222204
0.976766094995645 0.47353740650697
0.97691473504388 0.435574261360394
0.97691473504388 0.397624532519104
0.97691473504388 0.359743486044369
0.977105245528237 0.32197368705398
0.977105245528237 0.284119902843093
0.977433928451799 0.246296242343266
0.977812855898707 0.208565176486861
0.979301349902861 0.171310397646737
0.983212048636699 0.135011731481021
0.988253249145843 0.0993699571961411
0.988387234541435 0.0982187623045842
0.992364925973069 0.0648775196069138
0.993263046827896 0.0596086879988741
0.995729215515509 0.0450241957989573
0.996907868292356 0.0366528106065264
0.99751080257252 0.0317989488214125
0.997799708581765 0.0295472352248287
0.998281218597173 0.0244593541668493
0.998628743216989 0.0196073516681298
0.99888205935553 0.0156226315737893
0.999170965364775 0.00931923194844112
0.99956245394252 0.00521914994889487
1 1.09049055389329e-09
};
\addlegendentry{VL4AD}
\addplot [semithick, darkorange25512714]
table {%
0 1
0.00110747303543914 0.99994959970758
0.347960072352114 0.999533031225502
0.586979550478998 0.998464439902912
0.662610119247002 0.997079137408755
0.701953674549474 0.995378232771924
0.747914852281101 0.993070668611132
0.821292791585717 0.989893473129308
0.849812839150533 0.98527365133638
0.852932186641656 0.984593307365693
0.877409643598848 0.978549543726811
0.903417464996315 0.966750586421382
0.929900515843773 0.953915154921157
0.948474241307697 0.932609732801145
0.952879011187781 0.920266085278462
0.961424767200375 0.884705635417128
0.965643213639713 0.848591558768451
0.973441582367522 0.811697724668232
0.976596519729349 0.774461790595369
0.977450676626248 0.736998415088662
0.977942654250687 0.699389798235573
0.978344610437462 0.661721672222467
0.97875075366785 0.623940529949337
0.978914048368728 0.586202324651275
0.979236450726871 0.548473401608809
0.979391371340524 0.510722966459186
0.979531637301534 0.472981581290669
0.979682370871575 0.435250831676525
0.979799608092718 0.397532714304957
0.979799608092718 0.359785332528884
0.979921032357473 0.322000773748849
0.98004664366584 0.284178757708777
0.980161787365177 0.246553591571042
0.980536527768473 0.208848227486502
0.982054331077913 0.171567701073909
0.985644720975414 0.135428269429364
0.988761974944731 0.0994656662805847
0.989224643263884 0.0951874450156261
0.992946925035171 0.064938175962993
0.993891103369733 0.0582443087469745
0.996005560393917 0.0441248126241433
0.997064882427815 0.0358788251214856
0.997533831312387 0.0317920122109992
0.997881355932203 0.0288057060101435
0.998337743685938 0.0237313437635607
0.998637117304214 0.0188171527695337
0.99888205935553 0.015012431227
0.999196087626449 0.00859016049784357
0.999587576204194 0.00487586679400092
1 1.09049055389329e-09
};
\addlegendentry{VL4AD w. prompt}
\end{axis}

\end{tikzpicture}}
        \caption{\textbf{ID pixel retention rate on CityScapes as a function of  OOD recall (RA19/FS LaF)} VL4AD achieves a recall of 0.9 on both RA19 and FS LaF while correctly identifying at least $97\%$ of the CityScapes pixels as ID. Additionally, OOD prompting further enhances the ID retention rate on RA19.}
        \label{fig:cs}
\end{figure}
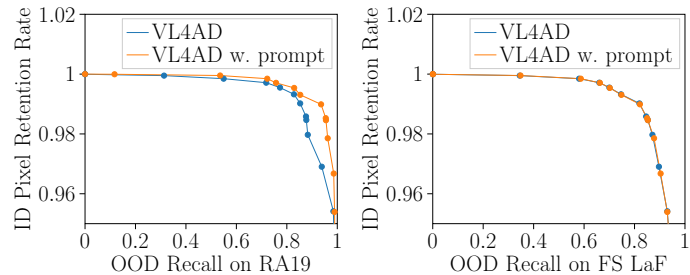
Compared to the original vision-only Mask2Former, VL4AD includes an additional text encoder, which might raise concerns about runtime efficiency. However, when the ID semantic classes and OOD prompts are determined and fed through the CLIP text encoder, we keep the resulting textual embeddings, thus avoiding further text encoder forward passes at test time. We compare VL4AD with the vision-only ConvNext \cite{Liu2022CVPR} with Mask2Former on an \verb|NVIDIA A100| and observe that VL4AD is approximately 5\% slower. This small difference is likely due to the fact that VL4AD's cosine similarities with text embeddings (see Eq.~\ref{eq:cos}) include more parameters than the linear layers in the vision-only Mask2Former. Additionally, we analyze the trade-off between anomaly detection and ID pixel identification, as shown in Fig.~\ref{fig:cs}. We apply uncertainty thresholds that were used in the evaluation of RA19 and FS LaF to CityScapes predictions to assess how many pixels are correctly classified (retained) as ID. The "void" class in CityScapes is excluded from the evaluation. We find that at a recall of 0.9 on RA19, approx. 2\% of the pixels in CityScapes are incorrectly identified as OOD. With OOD prompting, this figure is reduced to approximately 0.5\%. This demonstrates the effectiveness of OOD prompting. On FS LaF, both variants misclassify 3\% of CityScapes pixels at a recall of 0.9 since we use the OOD prompts from RA19.


\section{Limitations and Future Work} \label{limitations}
As discussed in Section \ref{pixel_eval}, VL4AD is currently limited by the constraints of CLIP's vision encoder, particularly when it comes to detecting small-sized obstacles on the road. This limitation is a root cause of VL4AD's weak pixel-level performance on the RO21 dataset. Enhancing CLIP's local representation capabilities for dense prediction tasks could be a promising direction for future work. Additionally, VL4AD relies on a concept dictionary and fixed prompt templates, as detailed in Appendix \ref{ensembling}. Although we experimented with learnable prompts \cite{Zhou2022IJCV}, the results did not yield improved anomaly detection performance. We speculate that this may be due to overfitting to the limited semantic classes in the CityScapes dataset. Developing effective training strategies for learnable prompts in the context of pixel-wise anomaly detection for limited classes remains an open challenge.


\section{Conclusion}
In this work, we present an approach to address the pixel-wise anomaly detection task. Most previous methods follow the vision-only paradigm and have limited knowledge of potential OOD object types due to the limited number of visual concepts in many standard training sets. As a consequence, they rely on outlier supervision with negative data to increase OOD sensitivity. Our method instead addresses the issue of limited OOD knowledge by leveraging vision-language models, which are exposed to a broader range of visual concepts during pre-training, making them more OOD-aware. To improve the alignment between ID textual embeddings and ID visual embeddings, we propose using max-logit prompt ensembling. Additionally, we introduce class merging to avoid high uncertainty on edge pixels between ID classes. We also propose a new scoring function tailored for vision-language models to enable data- and training-free outlier supervision through textual prompts. We implement and evaluate our approach on RA19, FS LaF, and the SMIYC benchmark and provide a comparison with existing methods such as RbA and UNO. Experimental results demonstrate that vision-language models are competitive to vision-only models in anomaly detection tasks but unlike the latter ones do not require additional data collection, labeling, and model retraining. Our strategy, which includes max-logit prompt ensembling, class merging, and OOD prompting, significantly improves the performance of our system and showcases the capability of vision-language models in detecting anomalies at the pixel level.


%
%
\clearpage
\bibliographystyle{splncs04}
\bibliography{main}

\newpage


\section{Appendix}
We provide details on the model architecture and hyperparameters in Section~\ref{training} and present further information on prompts and class merging in Section~\ref{ensembling} and~\ref{merging}, respectively. Next, we analyze issues with detecting small-size obstacles in Section \ref{obstacle} and, finally, present additional visualizations in \mbox{Section~\ref{fur_vis}.}


\subsection{Model Architecture and Hyperparameters}\label{training}
As mentioned previously, our model, VL4AD, is based on FC-CLIP \cite{Yu2023NIPS}.\footnote{\url{https://github.com/bytedance/fc-clip}} Unless specified otherwise, we use the same parameters as in the original study. In this section, we discuss central aspects of our model design and critical hyper\-parameters.


\subsubsection{In-Vocabulary and Out-of-Vocabulary Classifier} Eq.~\ref{eq:cos} computes the mask classification scores $\mathbf{c}_{i}$ using the cosine similarity between the processed visual embeddings $\mathbf{v}_i$ and the ID class textual embeddings $\mathbf{t}_{1}, \dots, \mathbf{t}_{K}$. The processed visual embeddings $\mathbf{v}_i$ represent a blend of visual features from the pixel-decoding part of the decoder $\mathcal{D}_\text{vis-lang}$, akin to those described in \cite{Ghiasi2022ECCV}, and those obtained directly via mask pooling from the encoder $\mathcal{E}_\text{vision, vis-lang}$.  Rather than simply summing these two types of visual features, FC-CLIP constructs a geometric ensemble of logits. This is achieved by computing the cosine similarity between the individual visual features and the ID class textual embeddings. FC-CLIP refers to these two sets of logits as the in-vocabulary and out-of-vocabulary classifiers, respectively \cite{Yu2023NIPS}. The final mask classification scores can then be written~as
\setcounter{equation}{3}
\begin{equation}
\label{equ:geometric_ensemble}
\mathbf{c}_{i}(j) = \begin{cases}
    (\mathbf{c}_{i, \text{in}}(j))^{(1-\alpha)} \cdot (\mathbf{c}_{i, \text{out}}(j))^\alpha, & \text{if}\ j \in K\\
    (\mathbf{c}_{i, \text{in}}(j))^{(1-\beta)} \cdot (\mathbf{c}_{i, \text{out}}(j))^\beta, & \text{otherwise}
\end{cases}
\end{equation}
where $\mathbf{c}_i(j)$ represent the $j$-th element of $\mathbf{c}_i$, $i=1, \dots, N$, and $N$ denotes the number of object queries. $\alpha, \beta \in [0, 1]$ balance the predictions between in- and out-of-vocabulary classifiers for ID and OOD prompting classes.\\\\
To rigorously examine the relationship between the features used for mask classification and OOD detection performance, we conduct a series of experiments for VL4AD without OOD prompting. We vary the parameter $\alpha$ from 0.0 to 1.0. Here, $\alpha = 0.0$ indicates that the model relies exclusively on visual features processed by the decoder, whereas $\alpha = 1.0$ implies that the model solely generates masks for CityScapes semantic classes, with classification scores entirely dependent on the raw visual features from the frozen vision encoder.\\\\
As illustrated in Tab. \ref{tab:alpha}, the AP remains consistently high across both the RA19 and the FS LaF dataset, significantly surpassing the vision-only variant shown in Tab. \ref{tab:abl}, regardless of the $\alpha$-value used. However, a noteworthy variation in the false positive rate on RA19 is observed. When using only the in-vocabulary classifier ($\alpha=0.0$), the model tends to overfit to the CityScapes training data, leading to suboptimal performance. In contrast, an exclusive reliance on the out-of-vocabulary classifier ($\alpha=1.0$) prevents the model from effectively learning from CityScapes data, resulting in a problematic uncertainty distribution and an FPR that is six times higher than in the optimal setup. The settings $\alpha=0.4$ and $\alpha=0.6$ demonstrate the best balance, indicating that moderate fine-tuning on in-distribution task data is essential for effective OOD detection. We also observe improved performance with OOD prompting at $\alpha=0.4$. Consequently, we use the same hyperparameters as FC-CLIP: $\alpha=0.4$ and~$\beta=0.8$. 
\setcounter{table}{4}
\begin{table}[tb]
    \caption{\textbf{Hyperparameter study for $\boldsymbol{\alpha}$} Pixel-wise metrics are reported on RA19 and FS LaF. Results are shown with the best in bold and the second best underlined.} %
\footnotesize
\centering
\adjustbox{max width=\textwidth}{%

\begin{tabular}{c c ccc ccc ccc ccc} 
 \toprule
 \multirow{2}{*}{$\alpha$} & \multirow{2}{*}{\shortstack{\rule{0pt}{10pt} OOD \\[0.3em] prompting}} 
 & \multicolumn{4}{c}{RA19} & \multicolumn{4}{c}{FS LaF} \\
 \cmidrule(r){3-6}  \cmidrule(r){7-10} 
&& \multicolumn{2}{c}{AP $\uparrow$} & \multicolumn{2}{c}{FPR $\downarrow$} &  \multicolumn{2}{c}{AP $\uparrow$} & \multicolumn{2}{c}{FPR $\downarrow$}  \\
 \midrule
 0.0  & \xmark & 87.4  & {\tiny ± 0.3}  & 14.0  & {\tiny ± 8.1}  & 73.7  & {\tiny ± 2.3}  & \textbf{3.5}  & {\tiny ± 0.2} \\
 0.2  & \xmark & 87.8  & {\tiny ± 0.5}  & 10.3  & {\tiny ± 4.2}  & 73.7  & {\tiny ± 2.3}  & \underline{3.7}  & {\tiny ± 0.3} \\
 0.4  & \xmark & \underline{88.9}  & {\tiny ± 0.5}  & \underline{6.4}  & {\tiny ± 2.1}  & 73.7  & {\tiny ± 2.3}  & 3.9  & {\tiny ± 0.2} \\
 0.6  & \xmark & \textbf{89.5}  & {\tiny ± 0.6}  & \textbf{5.3}  & {\tiny ± 0.7}  & \textbf{73.8}  & {\tiny ± 2.2}  & 4.0  & {\tiny ± 0.2} \\
 0.8  & \xmark & 86.9  & {\tiny ± 0.2}  & 28.1  & {\tiny ± 19.4} & \textbf{73.8}  & {\tiny ± 2.2}  & 4.3  & {\tiny ± 0.4} \\
 1.0  & \xmark & 85.5  & {\tiny ± 0.2}  & 33.6  & {\tiny ± 15.9} & 72.7  & {\tiny ± 2.8}  & 7.0  & {\tiny ± 1.1} \\
 \midrule
 0.4  & \cmark & \textbf{92.2}  & {\tiny ± 0.3}  & \textbf{5.8}  & {\tiny ± 1.9}  & \textbf{73.8}  & {\tiny ± 2.2}  & \textbf{4.6}  & {\tiny ± 0.5} \\
 0.6  & \cmark & 91.7  & {\tiny ± 0.3}  & 6.6   & {\tiny ± 2.1}  & \textbf{73.8}  & {\tiny ± 2.2}  & 4.9   & {\tiny ± 0.4} \\
 \bottomrule
\end{tabular}

    \label{tab:alpha}}
\end{table}


\subsubsection{Other Model Designs and Hyperparameters}
To mitigate the risk of overfitting, we take measures on the vision and text side of the model. We apply random cropping to the input images with a ratio of $1.1$ between original image size and crop. Following the methodology outlined in \cite{arandjelovic2023arxiv}, we construct a comprehensive list of textual prompts by considering all combinations of prompt templates and alternative concepts as detailed in Appendix~\ref{max-logit_concept}. For each generated prompt, we create prior to model training eight variants by applying dropout with a dropout rate of $1\%$ to the attention and feed-forward layers of the transformer-based text encoder. During the training process we sample from these eight variants. At the inference stage, we disable dropout and calculate the normalized mean of the textual embeddings with respect to the prompt templates provided in Appendix~\ref{prompt_template}. These mean embeddings serve as the representative textual embeddings for each alternative concept of a given semantic class.\\\\
We initially train our model on the COCO-panoptic dataset \cite{Lin2014ECCV} for 50 epochs using a batch size of 16, following the setup in \cite{Yu2023NIPS}. The learning rate is set to~$10^{-4}$. Subsequently, we fine-tune our model on the CityScapes dataset for 10K iterations with a batch size of 8 and a learning rate of $2.5\times10^{-5}$, maintaining a consistent setup for both VL4AD and the vision-only baselines. For the unfrozen vision-only variant, we reduce the learning rate of the encoder to 10\% of that of the decoder, as described in the Mask2Former paper. Across all experiments, we employ 250 object queries and 10 deformable attention layers, in line with the configurations used in \cite{Cheng2022CVPR, Yu2023NIPS}.


\subsection{Prompt Ensembling and OOD Prompting}\label{ensembling}%
In this section, we detail our approach to prompt ensembling and OOD prompting, including prompt templates and a concept dictionary for the in-data and out-of-data classes.


\subsubsection{Prompt Templates}\label{prompt_template}%
We adopt the prompt templates from \cite{Radford2021ICML, Yu2023NIPS}. Specifically, our list of templates is as follows:
\begin{quote}
    \setlength{\leftskip}{4mm}  
    \setlength{\rightskip}{0pt}  
    \textit{"A photo of a \{\}."\\
    "This is a photo of a \{\}."\\
    "There is a \{\} in the scene."\\
    "There is the \{\} in the scene."\\
    "A photo of a \{\} in the scene."\\
    "A photo of a small \{\}."\\
    "A photo of a medium \{\}."\\
    "A photo of a large \{\}."\\
    "This is a photo of a small \{\}."\\
    "This is a photo of a medium \{\}."\\
    "This is a photo of a large \{\}."\\
    "There is a small \{\} in the scene."\\
    "There is a medium \{\} in the scene."\\
    "There is a large \{\} in the scene."}
\end{quote}


\subsubsection{Concept Dictionary for Max-Logit Prompt Ensembling}\label{max-logit_concept}
\begin{table}[t]
    \caption{\textbf{Concept dictionary for the 19 CityScapes classes}}
    \centering
    \adjustbox{max width=\textwidth}{
    \begin{tabular}{|p{1.5cm}|p{2.7cm}|p{7.9cm}|} \hline  
          \textbf{Class ID}&\textbf{Single concept}& \textbf{Multiple alternative concepts}\\ \hline  
          0&road& road, railroad\\ \hline  
          1&sidewalk& sidewalk, pavement\\ \hline  
          2&building& building, buildings, edifice, edifices, house, ceiling\\ \hline  
          3&wall& wall, walls, brick wall, stone wall, tile wall, wood wall\\ \hline  
          4&fence& fence, fences\\ \hline  
          5&pole& pole, poles\\ \hline  
          6&traffic light& traffic light, traffic lights\\ \hline  
          7&traffic sign& traffic sign, stop sign\\ \hline  
          8&vegetation& vegetation, tree, trees, palm tree, bushes\\ \hline  
  9&terrain&terrain, river, sand, sea, snow, water, mountain, grass, dirt, rock\\ \hline  
  10&sky&sky, clouds\\ \hline  
  11&person&person, people\\ \hline  
  12&rider&motorcyclist, bicyclist, scooter rider, skateboarder, rollerblader, wheelchair user\\ \hline  
  13&car&car, cars\\ \hline  
  14&truck&truck, trucks\\ \hline  
  15&bus&bus, buses\\ \hline  
  16&train&train, trains, locomotive, locomotives, freight train, tram\\ \hline  
  17&motorcycle&motorcycle, motorcycles\\ \hline 
  18&bicycle&bicycle, bicycles, bike, bikes\\ \hline
    \end{tabular}}
    \label{tab:concepts}
\end{table}
We show our concept dictionary in Tab. \ref{tab:concepts}. The \textit{single concept} column simply lists the semantic classes. For the \textit{multiple alternative concepts} column, we primarily adopt the conceptual frameworks described in \cite{Ghiasi2022ECCV, Xu2023CVPR, Yu2023NIPS}, with adjustments based on the class definitions from CityScapes. For instance, we observe that the way the \textit{rider} class is conceptualized significantly influences the AP on the RoadAnomaly19 (RA19) dataset. This is likely because the generic description of \textit{rider} inadvertently correlates with animal anomalies found in RA19. By refining the concept of \textit{rider}, we reduce this correlation and mitigate related issues.


\subsubsection{OOD Prompting}\label{ood_prompt}
\begin{table}[t!]
\caption{\textbf{OOD prompting classes}}
    \centering
    \adjustbox{max width=\textwidth}{
    \begin{tabular}{|p{5cm}|p{5.7cm}|} \hline 
 \textbf{Dataset}&\textbf{OOD prompting classes}\\ \hline 
          \multirow{2}{*}{SMIYC (RA21, RO21)} & animal, animate being, dog, cat, horse, cow, sheep, zebra, giraffe, bird, elephant, carriage,  trailer, caravan, tractor\\ \hline 
 \multirow{2}{*}{RA19, FS LaF} & animal, animate being, dog, cat, horse, cow, sheep, zebra, giraffe, bird, elephant, cone, boulder, cardboard, tire\\ \hline 
          \multirow{2}{*}{RA19, FS LaF with RbA classes} & dining table, boat, banana, cow, tie, cake, pizza, sink, zebra, cat, toilet, keyboard, bear\\ \hline 
    \end{tabular}}
    \label{tab:ood_prompt}
\end{table}
The selection of the OOD classes is typically based on domain knowledge. For example, in real-world driving scenarios, traffic safety experts and statistical data from traffic administrations can help to compile lists of potential OOD objects. In our case, we extracted relevant information directly from the filenames within each dataset and list these OOD prompting classes in Tab.~\ref{tab:ood_prompt}. For the SMIYC benchmark, we derive the class names from the road anomaly track (RA21). For both RoadAnomaly19 (RA19) and FishyScapes Lost and Found (FS LaF), we adopt class names from RA19. Additionally, we consider names of 16 classes used in image-based outlier supervision of RbA, excluding all semantic classes that are clearly in-distribution, such as car, bicycle, and \mbox{stop sign.}


\subsection{Class Merging}\label{merging}
\begin{table}[t!]
\caption{\textbf{Class merging, from 19 to 8} We use the seven abstract superclasses defined in CityScapes plus a standalone class \textit{car}.}
    \centering
    \begin{tabular}{|p{2.3cm}|p{5.7cm}|} \hline 
         \textbf{Superclass}& \textbf{Semantic class}\\ \hline 
         flat& road, sidewalk\\ \hline 
         human& person, rider\\ \hline 
         car& car\\ \hline 
 other vehicle&truck, bus, train, motorcycle, bicycle\\ \hline 
         construction& building, wall, fence\\ \hline 
         object& pole, traffic sign, traffic light\\ \hline 
         nature& vegetation, terrain\\ \hline 
         sky& sky\\ \hline
    \end{tabular}
    \label{tab:8_class}
\end{table}
\begin{table}[t!]
    \caption{\textbf{Class merging, from 19 to 3} We construct three superclasses: static objects/background, moving objects, and humans.}
    \centering
    \begin{tabular}{|p{3cm}|p{5cm}|} \hline 
         \textbf{Superclass}& \textbf{Semantic class}\\ \hline 
         \multirow{2}{*}{\shortstack[1]{static objects/ \\ \hspace{-4mm}background}}
 & road, sidewalk, building, wall, fence, pole, traffic light, traffic sign, vegetation, terrain\\ \hline 
        moving objects& car, truck, bus, train, motorcycle, bicycle\\ \hline 
        human & person, rider\\ \hline
    \end{tabular}
    \label{tab:3_class}
\end{table}
Edge pixels between ID classes can lead to suboptimal OOD detection performance, especially as the number of ID classes grows. To address this, during testing, we concatenate textual prompts of related ID classes as alternative concepts within superclasses. We group the 19 semantic classes into eight superclasses, three superclasses, and a single superclass, respectively. The eight superclasses comprise the seven abstract superclasses defined in CityScapes plus a standalone class for \textit{car}. For the three superclasses, we group the 19 classes into static objects/background, moving objects, and humans. Details of our grouping are provided in Tab.~\ref{tab:8_class} and Tab.~\ref{tab:3_class}.\\\\
As indicated in Tab.~\ref{tab:num_class}, reducing the number of superclasses from 19 to 8 then to 3 leads to a monotonic improvement in both average precision (AP) and false positive rate (FPR) on the RA19 and FS LaF datasets. Notably, the FPR on both datasets decreases by more than $50\%$. When all semantic classes are merged into a single superclass, only one term remains on the right side in Eq.~\ref{eq:cos}, causing the softmax function to always output a probability of $1.0$. To address this, we switch to the sigmoid function. However, the sigmoid function also tends to saturate, which we speculate is why the three superclasses setup outperforms the single superclass setup on RA19.

\begin{table}
 \caption{\textbf{Impact of the superclass number (after merging) on VL4AD's performance on RA19 and FS LaF}} %
\footnotesize
\centering
\adjustbox{max width=\textwidth}{%
    \begin{tabular}{c c c c c c c c c} 
     \toprule
     \multirow{2}{*}{\vspace*{6mm}\shortstack{\rule{0pt}{8pt} \# of superclasses \\after merging}} & \multicolumn{4}{c}{RA19} & \multicolumn{4}{c}{FS LaF} \\
      \cmidrule(r){2-5} \cmidrule(l){6-9} 
      & \multicolumn{2}{c}{AP $\uparrow$}  & \multicolumn{2}{c}{FPR $\downarrow$}  & \multicolumn{2}{c}{AP $\uparrow$}  & \multicolumn{2}{c}{FPR $\downarrow$}  \\
     \midrule
       19 & 86.1 & \tiny{$\pm 0.3$} & 14.4 & \tiny{$\pm 2.7$} & 70.4 & \tiny{$\pm 2.8$} & 9.8 & \tiny{$\pm 0.6$} \\
       8  & 88.1 & \tiny{$\pm 0.5$} & 7.9  & \tiny{$\pm 0.8$} & 70.9 & \tiny{$\pm 2.8$} & 7.3 & \tiny{$\pm 0.9$} \\
       3  & \textbf{88.9} & \tiny{$\pm 0.5$} & \textbf{6.4} & \tiny{$\pm 2.1$} & 73.7 & \tiny{$\pm 2.3$} & 3.9 & \tiny{$\pm 0.2$} \\
       1  & 88.3 & \tiny{$\pm 0.7$} & 10.0 & \tiny{$\pm 3.3$} & \textbf{76.1} & \tiny{$\pm 1.7$} & \textbf{3.8} & \tiny{$\pm 0.9$} \\
     \bottomrule
    \end{tabular}}
    \label{tab:num_class}
\end{table}

\begin{figure}
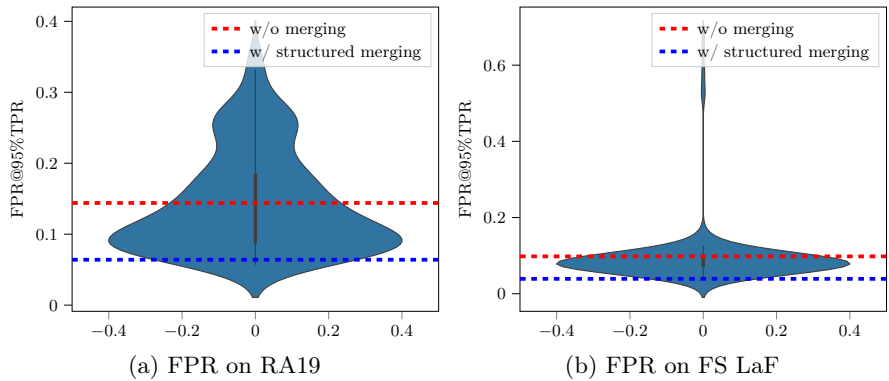

\begin{tabular}{cc}
\adjustbox{width=0.48\textwidth}{\input{images/fig6/RA19_FPR.tex}}&
\hfill
\adjustbox{width=0.48\textwidth}{\input{images/fig6/FS_FPR.tex}}\\
(a) FPR on RA19&(b) FPR on FS LaF
\end{tabular}
\caption{\textbf{Structured merging vs. random merging vs. no merging} The dashed red lines indicate the average false positive rate (FPR) observed without class merging. The dashed blue lines represent the average FPR when semantic classes are merged into three superclasses based on their semantic similarities. The violin plots summarize the FPRs from 300 random merging runs.}
\label{fig:merging}
\end{figure}

To better understand the effects of different merging strategies, we conduct additional experiments. These involve randomly merging 19 semantic classes into three superclasses, a process we repeat 100 times for all three models, each trained with different random seeds. We determine the FPR on RA19 and FS LaF and illustrate the results in Fig.~\ref{fig:merging}. This approach allows us to compare random merging against structured merging, which utilizes semantic similarities between classes. The findings indicate that employing any form of class merging generally results in a lower mean FPR than using no merging. Notably, structured merging (shown as a dashed blue line) consistently outperforms both random merging and the absence of merging (shown as a dashed red line). In all experiments in the main part of this work, we utilize structured merging with three superclasses.


\subsection{Issues for Small-Size Obstacles Detection}\label{obstacle}
\begin{figure}
\centering
\adjustbox{width=0.97\textwidth}{\includegraphics[]{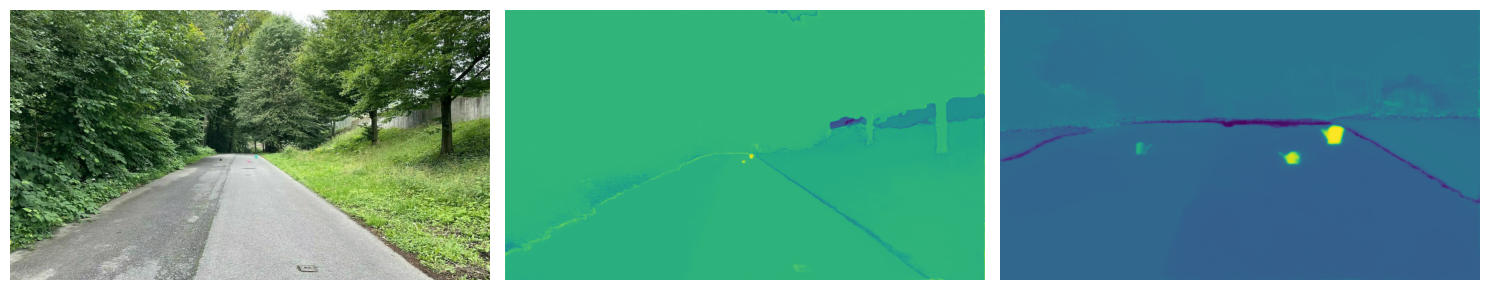}}\\
\adjustbox{width=0.97\textwidth}{\includegraphics[]{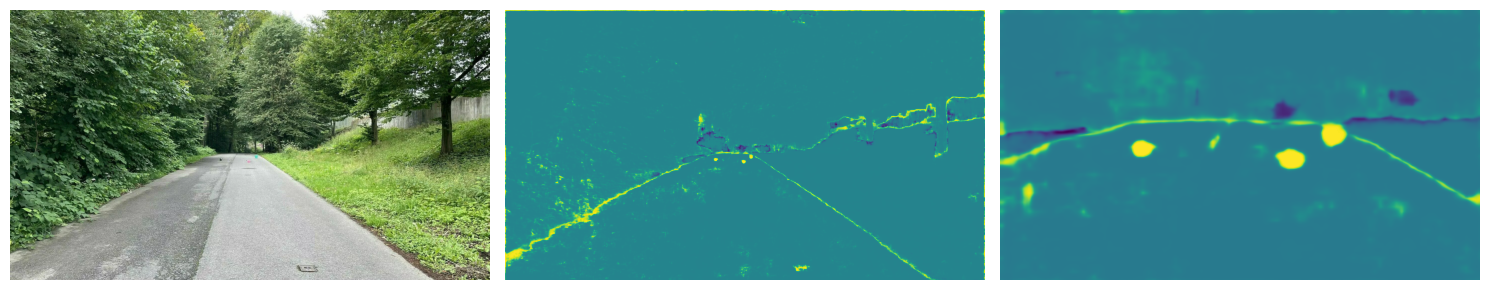}}\\
\adjustbox{width=0.97\textwidth}{\includegraphics[]{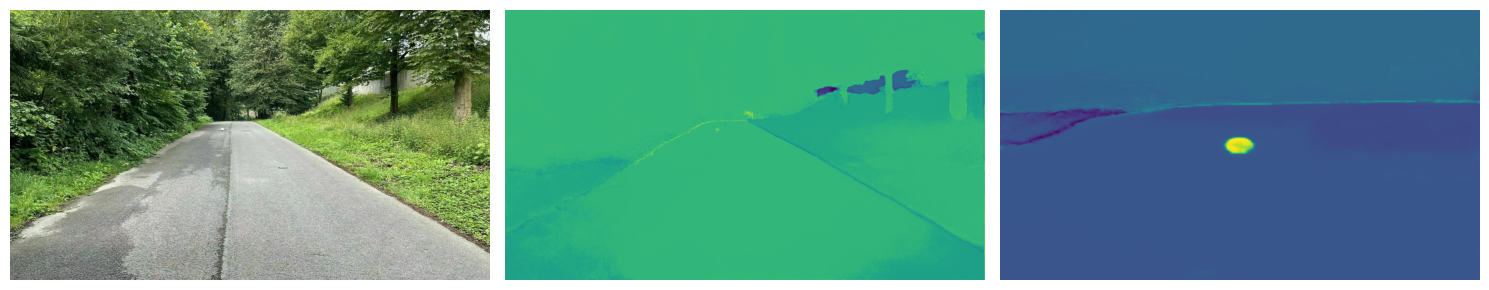}}\\
\adjustbox{width=0.97\textwidth}{\includegraphics[]{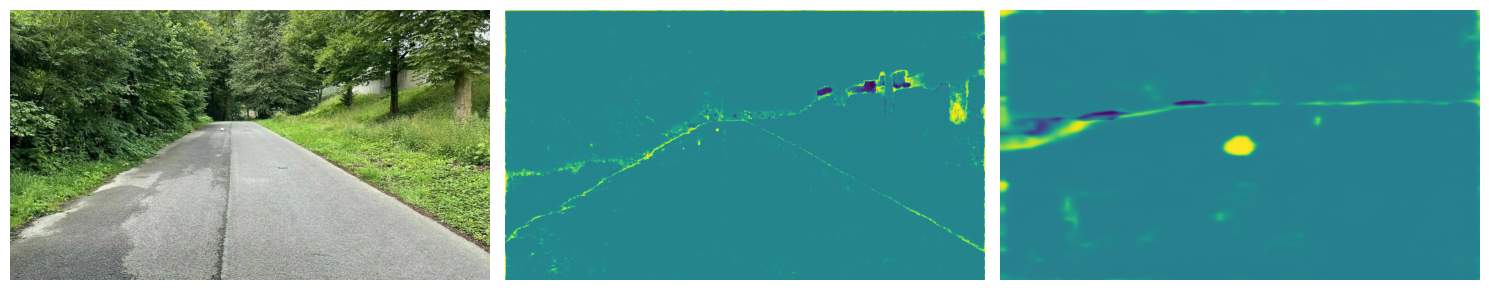}}\\
\adjustbox{width=0.97\textwidth}{\includegraphics[]{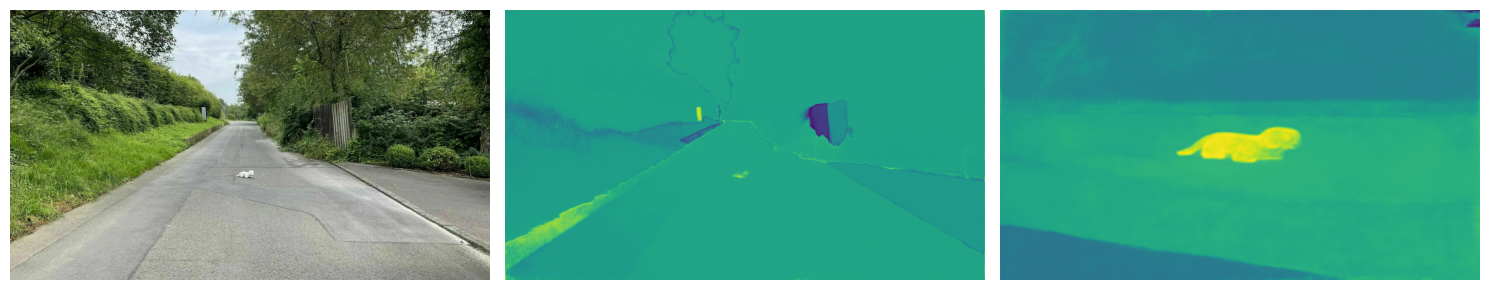}}\\
\adjustbox{width=0.97\textwidth}{\includegraphics[]{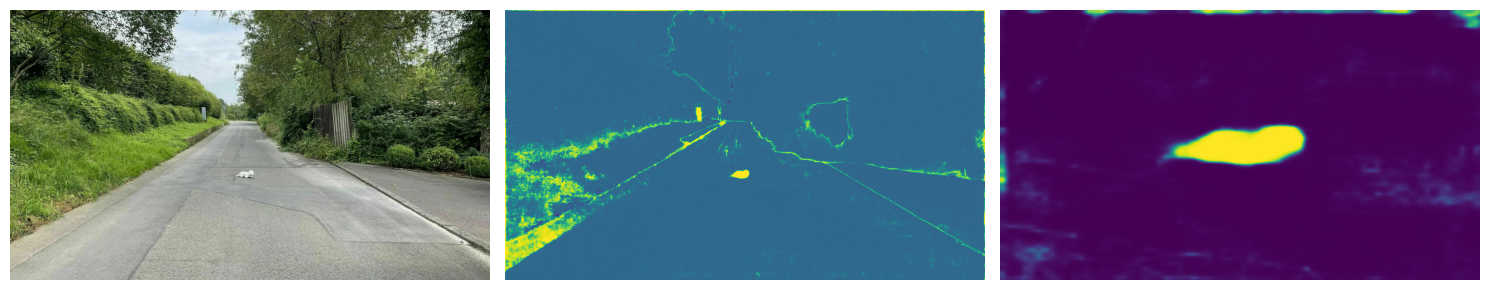}}\\
\adjustbox{width=0.97\textwidth}{\includegraphics[]{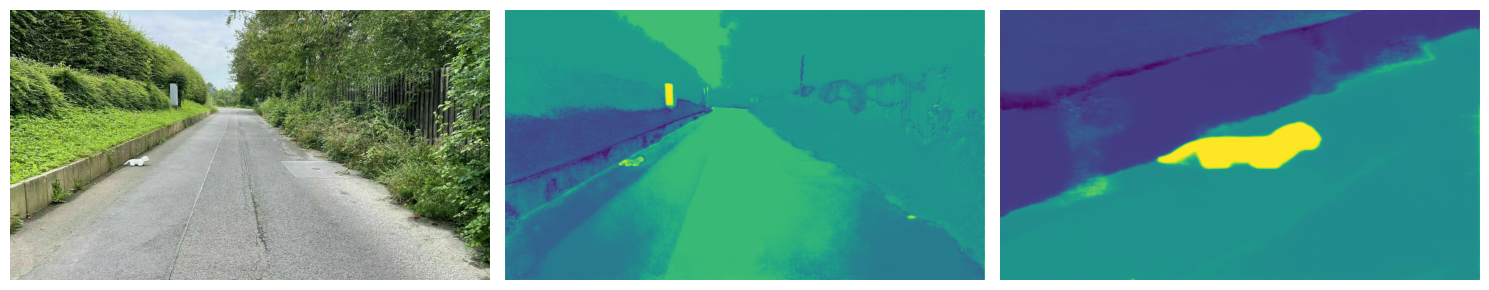}}\\
\adjustbox{width=0.97\textwidth}{\includegraphics[]{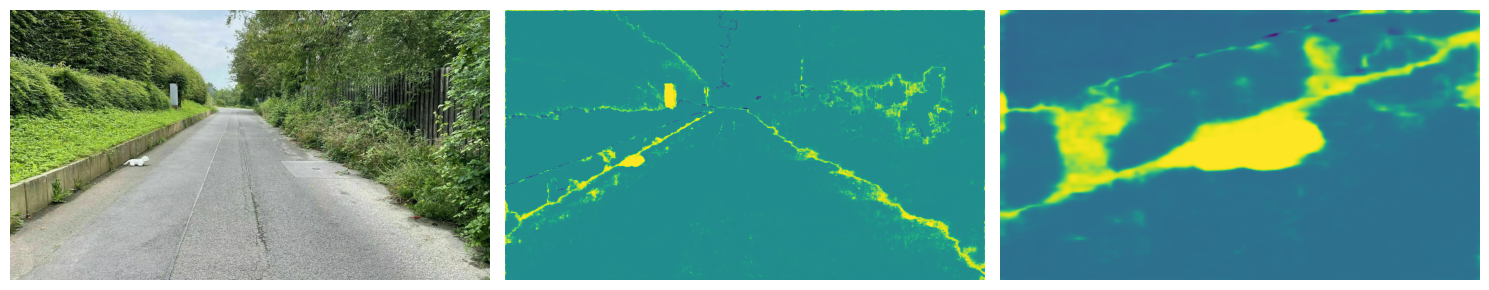}}\\
\caption{\textbf{VL4AD (ours, odd rows) and RbA (even rows) on the SMIYC \mbox{obstacle} track} The input images are shown on the left, followed by the respective uncertainty scores in the middle, and the uncertainty scores with cropped-out ground truth area as input on the right.}
\label{fig:obs}
\end{figure}

Although VL4AD shows strong performance on the anomaly track (RA21) of SMIYC, as well as on RoadAnomaly19 and FishyScapes Lost and Found (FS LaF), it does not perform as well on the obstacle track (RO21) of SMIYC. A key difference between the tracks are the sizes of the anomalous objects, which tend to be much smaller in the obstacle track. This performance issue may be attributed to the characteristics of the frozen CLIP encoders, which were pre-trained on object-centered text-image pairs. Implementing geometric image-based data augmentations or extracting visual features from the model's earlier layers might mitigate this issue. In Fig.~\ref{fig:obs}, we present some visualizations from the SMIYC obstacle track validation set to illustrate these observations.
For VL4AD, the differences between uncertainty scores for obstacles and the road are not particularly strong, especially when the obstacles are at a distance. However, when cropping the area around an obstacle (which can be done if respective ground truth information is available) and feeding only this crop to the network, this contrast is significantly enhanced without a substantial increase in false positive pixels. In contrast, while RbA consistently shows a good contrast between obstacles and the road, it tends to introduce more false positives when using the cropped-out input.


\subsection{Further Visualizations for the SMIYC Anomaly Track and FishyScapes Lost and Found} \label{fur_vis}
\begin{figure}
\centering
\adjustbox{width=0.99\textwidth}{\includegraphics[]{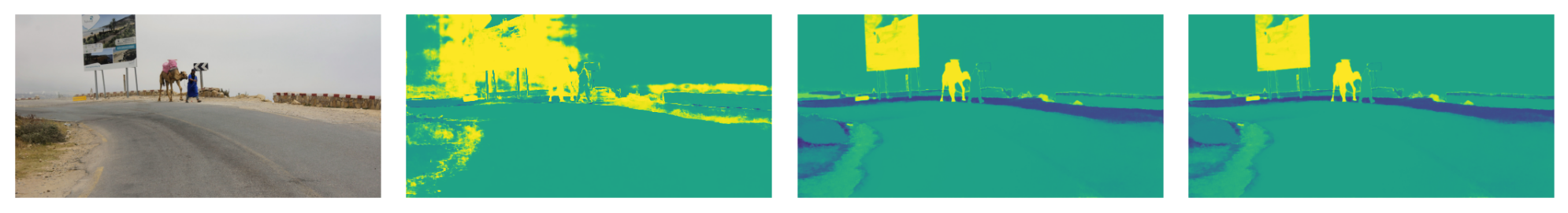}}\\
\adjustbox{width=0.99\textwidth}{\includegraphics[]{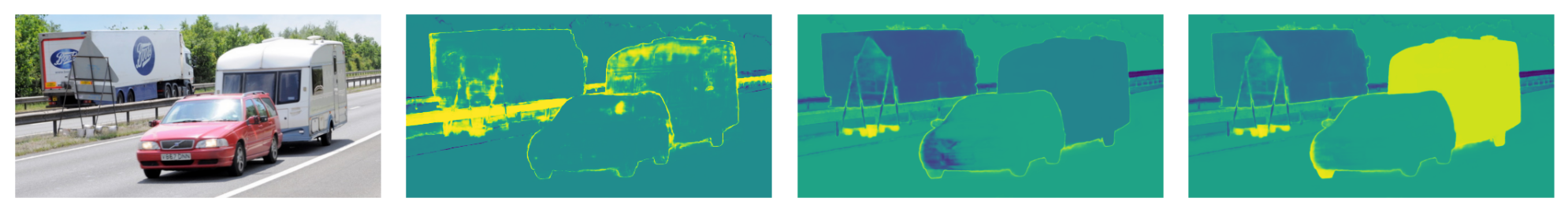}}\\
\adjustbox{width=0.99\textwidth}{\includegraphics[]{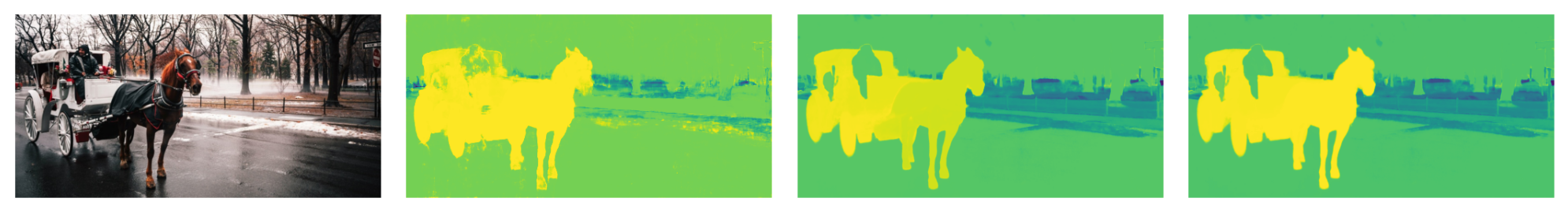}}\\
\adjustbox{width=0.99\textwidth}{\includegraphics[]{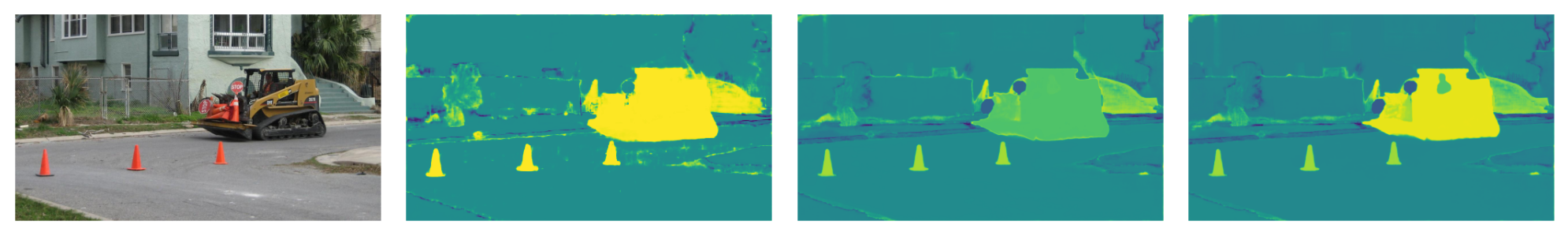}}\\
\adjustbox{width=0.99\textwidth}{\includegraphics[]{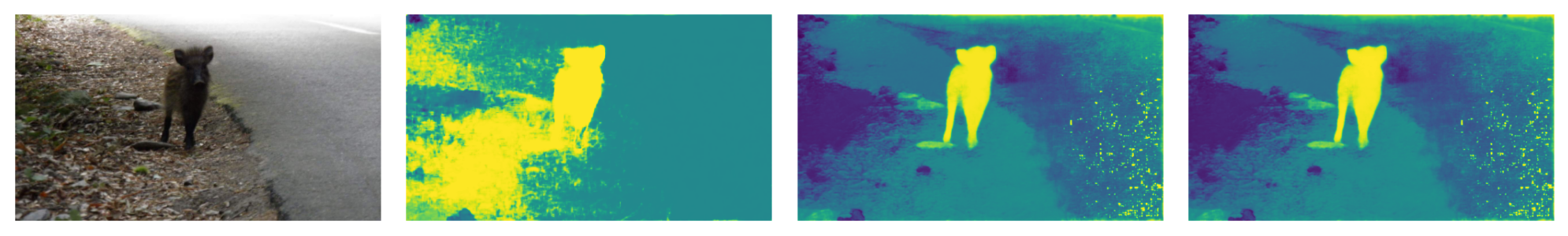}}\\
\adjustbox{width=0.99\textwidth}{\includegraphics[]{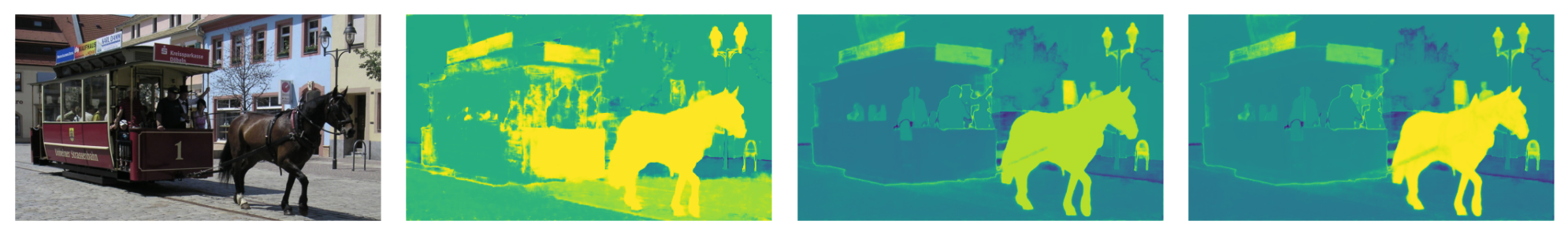}}\\
\adjustbox{width=0.99\textwidth}{\includegraphics[]{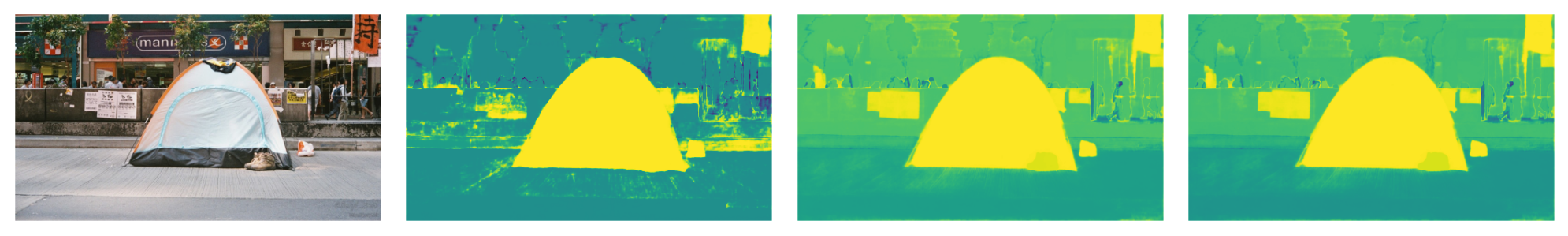}}\\
\adjustbox{width=0.99\textwidth}{\includegraphics[]{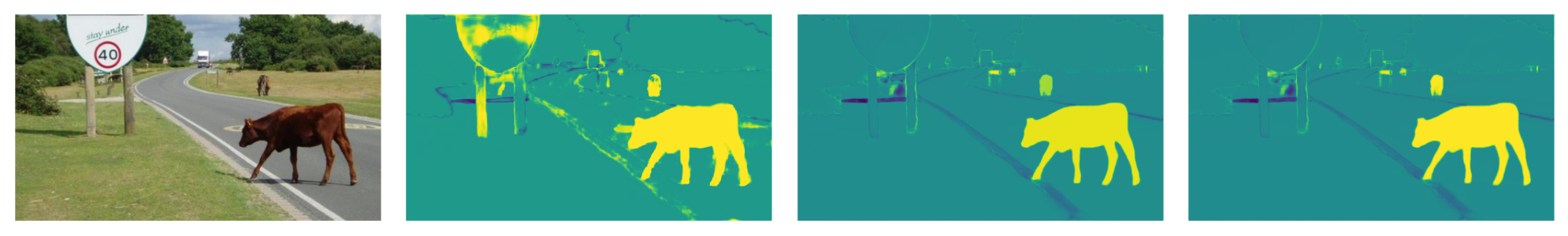}}\\
\caption{\textbf{VL4AD (ours, 3rd/4th column) and RbA (2nd column) on the SMIYC anomaly track} From left to right, the sequence is as follows: the input image, RbA with outlier supervision, VL4AD, and VL4AD with OOD prompting. VL4AD offers a cleaner and more coherent detection output. Furthermore, the benefits of OOD prompting become apparent, see the caravan depicted in the second row.}
\label{fig:ano}
\end{figure}

\begin{figure}
\centering
\adjustbox{width=0.98\textwidth}{\includegraphics[]{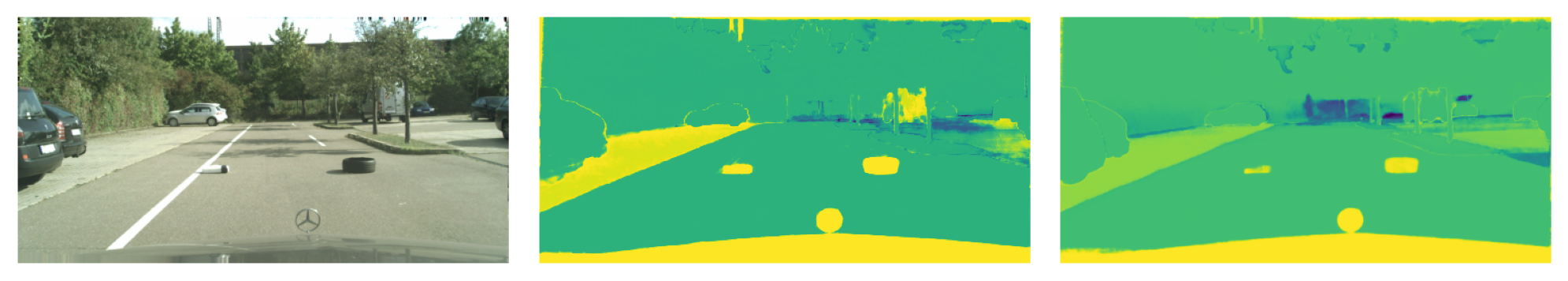}}\\
\adjustbox{width=0.98\textwidth}{\includegraphics[]{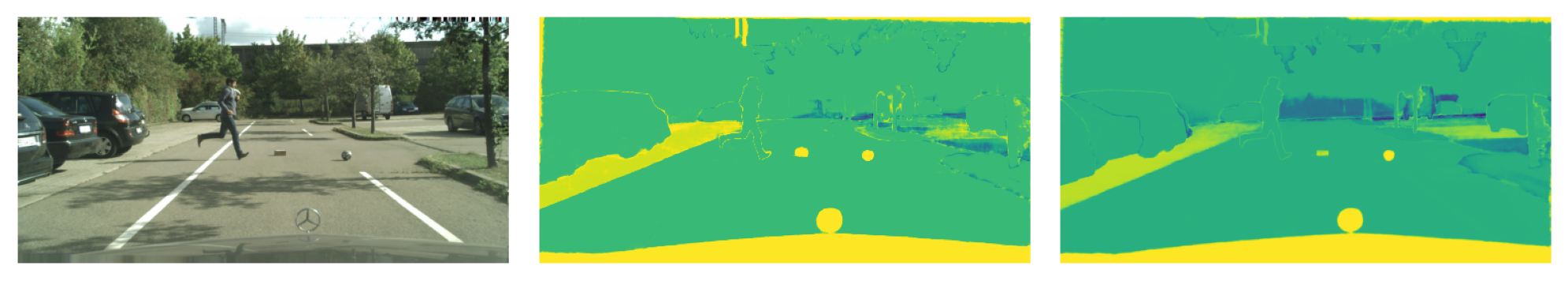}}\\
\adjustbox{width=0.98\textwidth}{\includegraphics[]{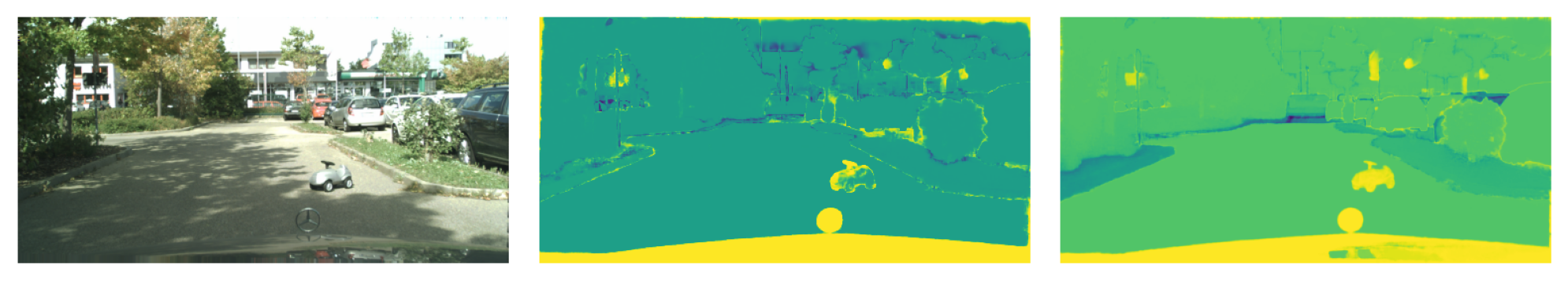}}\\
\adjustbox{width=0.98\textwidth}{\includegraphics[]{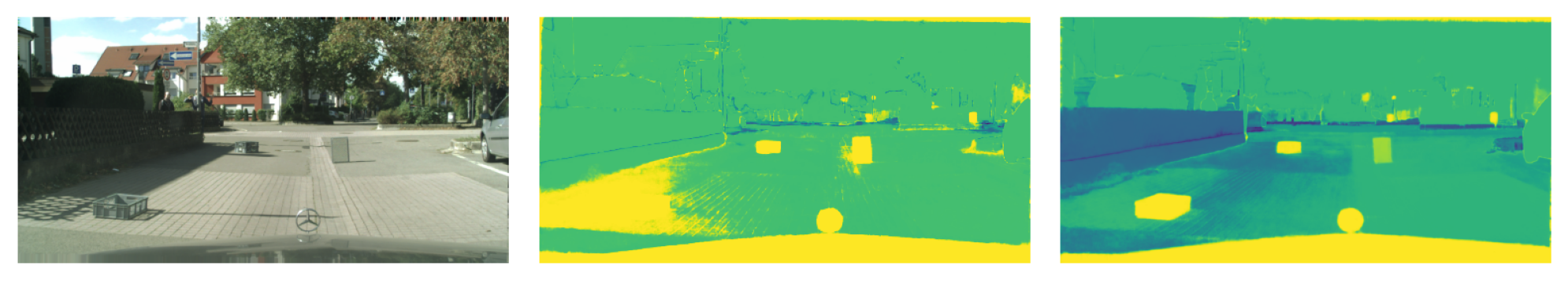}}\\
\adjustbox{width=0.98\textwidth}{\includegraphics[]{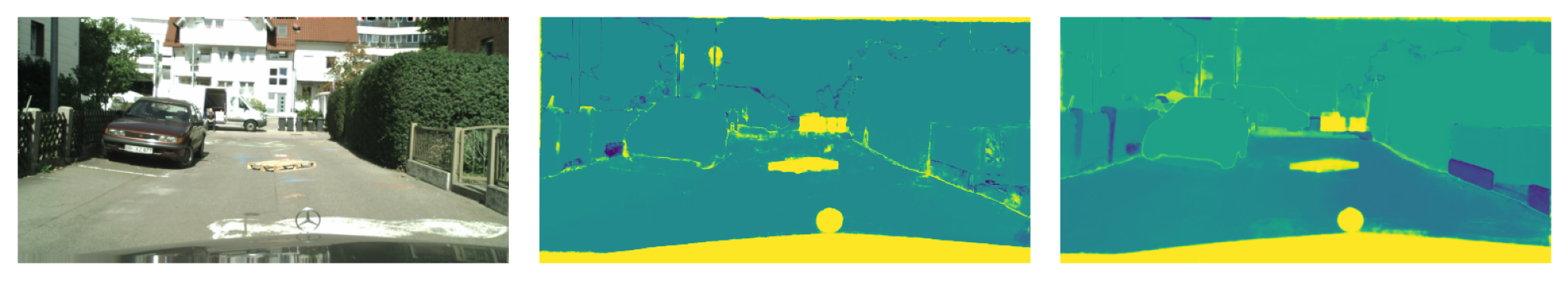}}\\
\adjustbox{width=0.98\textwidth}{\includegraphics[]{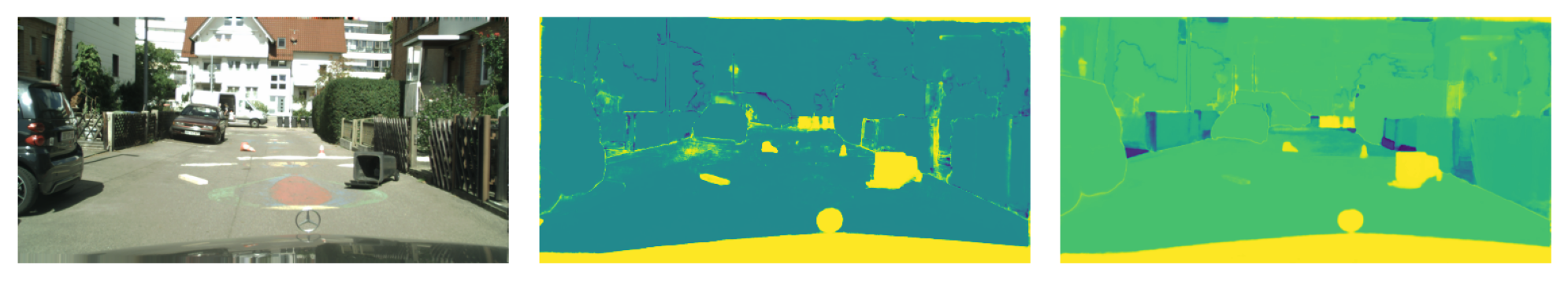}}\\
\adjustbox{width=0.98\textwidth}{\includegraphics[]{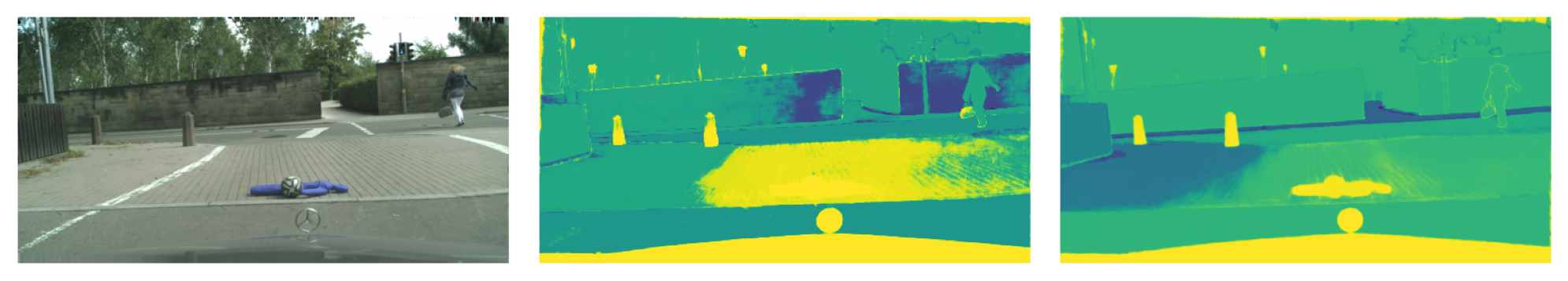}}\\
\adjustbox{width=0.98\textwidth}{\includegraphics[]{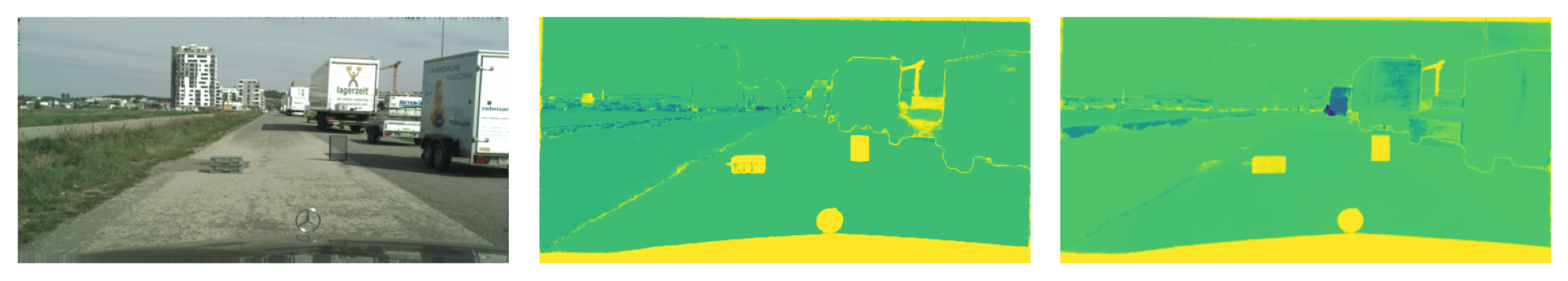}}\\
\caption{\textbf{VL4AD (ours, right column) and RbA (middle column) on FS LaF} From left to right, the sequence is as follows: input image, RbA with outlier supervision, and VL4AD. VL4AD demonstrates slight advantages over RbA, particularly evident in its prediction of fewer false positive pixels, as can be seen in the fourth and \mbox{seventh row.}}
\label{fig:ano1}
\end{figure}

We present further visualizations from the anomaly track (RA21) of the SMIYC benchmark and FishyScapes Lost and Found (FS LaF) in Fig.~\ref{fig:ano} and Fig.~\ref{fig:ano1}, respectively. These illustrations provide deeper insights into the detection capabilities of our models across various challenging scenarios. 
\end{document}